# A Review on Viewpoints and Path-planning for UAV-based 3D Reconstruction

Mehdi Maboudi[1*], MohammadReza Homaei[2], Soohwan Song[3], Shirin Malihi[4,5], Mohammad Saadatseresht[2], and Markus Gerke[1]

[1]*Institute of Geodesy and Photogrammetry, Technische Universität Braunschweig, Germany*

[2]*School of Surveying and Geospatial Eng., College of Eng., University of Tehran, Tehran 1439957131, Iran*

[3]*Intelligent Robotics Research Division, ETRI, Daejeon 34129, Republic of Korea*

[4]*Tandon School of Engineering, New York University, New York, NY 11201, USA*

[5]*School of Engineering, Institute for Infrastructure and Environment, University of Edinburgh, Edinburgh, UK*

**ABSTRACT**:

Unmanned aerial vehicles (UAVs) are widely used platforms to carry data capturing sensors for various applications. The reason for this success can be found in many aspects: the high maneuverability of the UAVs, the capability of performing autonomous data acquisition, flying at different heights, and the possibility to reach almost any vantage point. The selection of appropriate viewpoints and planning the optimum trajectories of UAVs is an emerging topic that aims at increasing the automation, efficiency and reliability of the data capturing process to achieve a dataset with desired quality. On the other hand, 3D reconstruction using the data captured by UAVs is also attracting attention in research and industry. This review paper investigates a wide range of model-free and model-based algorithms for viewpoint and path planning for 3D reconstruction of large-scale objects. The analyzed approaches are limited to those that employ a single-UAV as a data capturing platform for outdoor 3D reconstruction purposes. In addition to discussing the evaluation strategies, this paper also highlights the innovations and limitations of the investigated approaches. It concludes with a critical analysis of the existing challenges and future research perspectives.

*Keywords*: Viewpoint planning, Path planning, UAV, Drone, 3D reconstruction, Aerial photogrammetry, Active vision, View-path planning

## 1 Introduction

A complete dataset which meets the predefined quality measures is the fundamental input for various applications like 3D reconstruction of large-scale complex objects with a high level of details (for example cultural heritage documentation), inspection of construction sites, and Building Information Modeling (BIM) just to name a few. Many sensors and platforms for capturing high-quality images and point clouds have been introduced in the last decades. Especially, unmanned aerial vehicles (UAVs) are widely used as platforms to carry data capturing sensors. More specifically, multi-copters (from this point forward we use the term UAV to refer to multi-copter) are agile, capable of performing autonomous data acquisition, can fly at different heights, and can reach almost any vantage point (Song et al., 2021). Hence, UAV-based 3D reconstruction is still attracting attention as a promising approach for generating semantic and geometric information about man-made and natural objects.

UAVs are utilized in various monitoring and inspection projects, for example, in agriculture (González-Jorge et al., 2017; Radoglou-Grammatikis et al., 2020), forestry (Torresan et al., 2016), environmental sciences (Eltner et al., 2022; Eskandari et al., 2020), Hydrology (Acharya et al., 2021), industrial inspection of large structures (Maboudi et al., 2021; Seo et al., 2018; Sun and Ma, 2022), Virtual Reality (Sharma et al., 2019) and civil engineering (Debus and Rodehorst, 2021; Jacob-Loyola et al., 2021; Rakha and Gorodetsky, 2018). Especially the demand for high-quality 3D models in many applications such as construction (Li and Liu, 2018), urban modelling (Malihi et al., 2018; Song et al., 2021), aerial cinematography (Bucker et al., 2020), transportation (Sahebdivani et al., 2020; Xu and Turkan, 2020), landscape planning, archeology (Fiz et al., 2022), emergency rescue, and disaster analysis (Kerle et al., 2019) is still increasing. Nevertheless, restricted access to the objects

---

and practical considerations like limited flight time, safety protocols, battery constraints, and legal regulations should also be considered. Furthermore, 3D reconstruction of complex objects with high level of details necessitate a thorough viewpoint and flight path planning. The quality of such 3D reconstructions highly depends on the acquisition process and camera network configuration (Koch et al., 2019; Zhou et al., 2020a).

Although it is possible to generate 3D reconstruction products with impressive quality using state-of-the-art algorithms, the success of these reconstruction methods depends to a large degree on the input images with predefined specifications (Hepp et al., 2018b; Hoppe et al., 2012; Saponaro et al., 2019; Schmid et al., 2012; Seitz et al., 2006; Song et al., 2020a). Furthermore, adding more images cannot always deliver better reconstruction, and even diminishing returns at some points are reported in the literature (Hepp, 2018; Hornung et al., 2008; Seitz et al., 2006). Moreover, due to the limited battery capacity of the current UAVs and possible restrictions in accessing the area of interest, data capturing should be performed in a limited time (Smith et al., 2018). Hence, efficient viewpoint and path planning algorithms are required to capture the data for the successful reconstruction of 3D objects with various sizes, shapes, and complexities (Kaba et al., 2017; Kuang et al., 2020; Roberts, 2019; Song et al., 2020b).

The selection of appropriate viewpoints and trajectory for 3D reconstruction has been a research topic for some decades (Ahmadabadian et al., 2014; Aloimonos et al., 1988; Connolly, 1985; Debus and Rodehorst, 2021; Peng and Isler, 2017; Scott et al., 2003; Song et al., 2021). In robotics, view and path planning is a research focus of many studies on 3D reconstruction to make the robots explore and reconstruct unknown environments more competently (Chen et al., 2011; Scaramuzza et al., 2014; Scott et al., 2003). This task is known as the active vision or view-path-planning problem in robotics (Roberts et al., 2017; Song et al., 2021). In Photogrammetry, the problem of determining the optimal viewpoints for accurate measurement of 3D scenes or objects is called camera network design (Fraser, 1984; Hoppe et al., 2012; Mason, 1997; Saadatseresht et al., 2005) where mainly the geometric aspects of the problem are investigated in controlled environments for highly accurate measurement of artificial targets using convergent imaging.

*1.1 Scope of the paper:*

The review includes peer-reviewed publications and theses, focusing on wide range of model-free and model-based algorithms on viewpoint and path planning for 3D reconstruction of large-scale objects. In addition to discussing the evaluations strategies, this paper also highlights the innovations and limitations of the investigated approaches. It concludes with a critical analysis of the existing challenges and future research perspectives. Since various platforms and sensors are employed for different 3D reconstruction applications, we exclude the studies which are not within the scope of this paper.

**Platform**: Many researchers in robotics and other communities investigate view and path planning for moving robots, UGVs (Unmanned Ground Vehicles) and wheeled robots (Cao et al., 2021; Hosseininaveh and Remondino, 2021; Huang et al., 2020), Autonomous Underwater vehicles (Panda et al., 2020), robotic arms (Bogaerts et al., 2019; Wu et al., 2014) as well as multi-UAVs (Jing et al., 2020; Nagasawa et al., 2021; Sadeghi et al., 2019) for optimal data capturing. Although there are some common issues between path planning for these platforms and UAVs, the assumptions, challenges, trajectories, platforms abilities, and environmental conditions could be very different. For example, in contrast to UGVs, UAVs should consider a 6D search space for the sensor poses within the reconstruction workflow (Schmid et al., 2012). Therefore, this paper focuses on viewpoint and path planning for a single UAV.

**Sensor and 3D** representation: Various sensors are installed on UAVs for data capturing in different applications (Nex et al., 2022). Monocular RGB camera (Song et al., 2020a), RGB-D sensor (Hardouin et al., 2020), depth cameras (Deris et al., 2017; Xu et al., 2021), structured light scanners (Ułanowicz and Sabak, 2021), multi-spectral camera (DadrasJavan et al., 2019), stereo cameras and LiDAR (Alsadik and Remondino, 2020; Bolourian and Hammad, 2020; Yoder and Scherer, 2016) are the most commonly used sensors (Yan et al., 2021). Compared to single RGB cameras, other sensors entail special hardware and setup, making them very efficient for some applications. However, it usually comes at the cost of more technical requirements, bounds applications and makes them more expensive. Different 3D representation methods have been developed depending on the type of scanning sensors. Images acquired from a monocular camera are generally processed by MVS algorithms (Schönberger et al., 2016) to reconstruct 3D models. MVS generates a dense 3D reconstruction offline by matching the stereo correspondences of all images. This method can estimate a wide range of depths based on various baseline distances of images; therefore, MVS is appropriate for large-scale modeling. RGB-D sensors and stereo cameras can estimate accurate and dense depth maps in real-time. The estimated depth maps can be integrated into a volumetric model or a surface model by the volumetric mapping (Hornung et al., 2013) or TSDF mapping (Whelan et al., 2014), respectively. Volumetric models represent a coarse 3D shape based on occupancy probability, whereas surface models represent the precise surfaces of an object. LiDAR can acquire very accurate point cloud data in real-time. The point clouds are relatively sparse compared to the depth maps acquired from depth sensors and cannot represent dense surfaces. Therefore, LiDAR-based methods (Qin et al., 2019; Yoder and Scherer, 2016) generally employ the volumetric mapping method for scanning.

**Application**: Geometric criteria like completeness and accuracy of the 3D models acquired from UAV missions are the most interesting properties discussed in the literature (Roberts et al., 2017). However, researchers also aimed at other properties like semantic information, which could be extracted from the UAV data (Popovic et al., 2017; Stache et al., 2021; Valente et al., 2013). Apart from many other applications like precision agriculture (Basiri et al., 2022; Just et al., 2020), indoor exploration and reconstruction (Shen et al., 2012; Sun et al., 2021; Zhu et al., 2018) to name a few, we focus on viewpoint and path planning for 3D reconstruction of complex outdoor man-made objects like buildings and bridges which is a topic of interest for many applications (Almadhoun et al., 2019). A closely related topic is coverage path planning (Bircher et al., 2015; Galceran and Carreras, 2013; Kaba et al., 2017; Peng and Isler, 2020; Shang et al., 2020; Tan et al., 2021) which deals with optimizing the trajectories that cover a known 2D or 3D environment. This topic could be considered as a set-cover problem (Jing et al., 2019) or an art gallery problem followed by the shortest path optimization (Almadhoun et al., 2016; Heng et al., 2015; Papachristos et al., 2019a). However, since these approaches do not consider MVS heuristics and 3D reconstruction requirements and the optimal coverage path with low overlapped rate is desired, they are out of the scope of this paper.

*1.2 SfM and MVS heuristics in viewpoint planning*

Since many of the 3D reconstruction approaches rely on structure from motion (SfM) and multi-view stereo (MVS), we shortly list the most important heuristics as follows (see also Figure 1a):

(1) Distance: the distance between camera viewpoints and object surface defines the resulting model resolution and depends on the desired GSD and the camera intrinsics (Koch et al., 2019; Peng and Isler, 2018).
(2) Multiple views: every part of the scene has to be observed from multiple views (theoretically at least two views and practically more for increasing the reliability and accuracy of 3D reconstruction) from different perspectives with sufficient overlap between the views. Matching the corresponding points in overlapping images is the prerequisite for robust camera poses estimation and for triangulating 3D object points (Hoppe et al., 2012; Mostegel et al., 2016; Roberts et al., 2017; Schmid et al., 2012).
(3) Observation and parallax angle: Shallow observation angles between the optical-axis of the cameras and surface normals are preferred in MVS. Moreover, large parallax angles (a.k.a angular separation) between cameras (also known as the B/H ratio in photogrammetry) are favored to increase the triangulation quality. However, it makes it more difficult to find correspondences (matching) between the images, especially for complex 3D structures (Koch, 2020; Mostegel et al., 2016; Peng and Isler, 2019; Smith et al., 2018).

Various models are used in the literature to control these parameters (Koch, 2020; Peng and Isler, 2018; Roberts, 2019). Three examples are illustrated in Figure 1b-d.

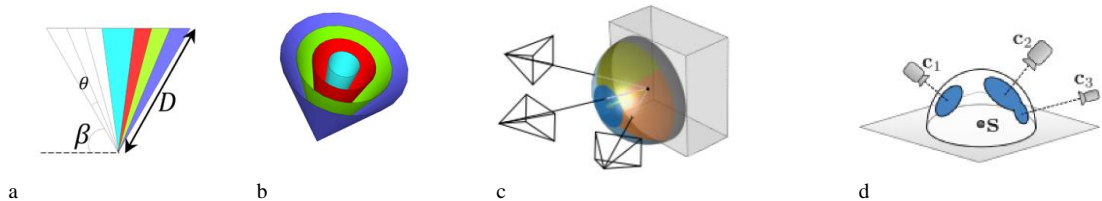

a  b  c  d

Figure 1: SfM and MVS heuristics in viewpoint planning: a) Parameters, b) co-apex cones for forcing the suitable cameras to adhere to the heuristics, c) considering the segments of a hemisphere for viewpoint planning (Koch, 2020), and d) MVS-aware coverage model (Roberts, 2019)

The parameters defined above are visualized in Figure 1a. The strategy which is used in (Koch, 2020) is illustrated in Figure 1b, where a hemisphere is considered around the sample point. While the hemisphere is divided into four segments, the ideal camera constellations intersect the hemisphere in different segments. The coverage model introduced by (Roberts, 2019) is depicted in Figure 1c. Since each camera covers a disk area on the hemisphere around the sample point, the usefulness of the cameras is measured by total solid angle covered by all the disks.

*1.3 Problem formulation*

Since different terms, objectives, priorities and approaches are utilized for viewpoints and path planning in various communities, we formulate the problem again to provide a holistic definition that covers different strategies and nomenclature, consistently.

Determining "good", "ideal" or "reasonable" imaging positions is a common research topic in various communities such as robotics, computer graphics, computer vision, and photogrammetry (Hoppe et al., 2012). In UAV-based research, some viewpoint and path planning approaches (Hoppe et al., 2012; Schmid et al., 2012) solve the planning problem in two steps: The first step solves the optimum viewpoint while ignoring the travel budget and the second step, minimizes the cost of visiting all viewpoints which are computed in the first step. However, some other approaches solve both viewpoint and path planning problems in a unified approach (Hepp et al., 2018b; Roberts et al., 2017). This is mainly a domain specific concern and depends on the application and requirements/priorities. Giving priority to the reconstruction of the object is useful (usually leads to) first solving the camera poses (sensor placement, a.k.a viewpoint selection) and then finding the shortest/fastest trajectory to visit all those points. The constraints are defined on the object and quality has the highest priority. In this case, after computing the vantage viewpoints, TSP (shortest Hamiltonian path between the viewpoints) is a commonly used approach in the literature. On the other hand, giving priority to resources (flight time, trajectory length, or the number of the images) the constraints are defined on the sensor (also constrained by the obstacles). The aim would be to get as much as possible but hardly constrained by the resources. For maximum gain, the classical knapsack problem could be used. However, costs are not fixed in view-path planning (depending on the distance to the previous viewpoint). Therefore, this is a combination of knapsack and TSP which is called the orienteering problem (Gunawan et al., 2016; Vansteenwegen et al., 2011). The orienteering problem (reward-collecting graph optimization problem) was first introduced by (Golden et al., 1987). The objective is maximizing the total gain collected from visited nodes. Because of the limited time budget, all available nodes may not be selected. In orienteering problem marginal rewards are additive, but in MVS the usability of each camera is not additive and depends on some heuristics which could be defined considering the poses of other cameras. Different strategies are employed to tackle this problem (Hepp et al., 2018b; Roberts et al., 2017). Generally, we formulate the viewpoint and path planning as follows:

Considering a camera $K$ with fixed intrinsics (at least during the data capturing) mounted on a drone, the objective of viewpoint and path planning is to find a feasible trajectory $T$ of the drone which meets some constraints $\Omega$ and passes through optimal camera positions $P = \{p_1, p_2, ..., p_N\}$ for taking high-quality pictures at optimal orientations $\Phi = \{\varphi_1, \varphi_2, ..., \varphi_N\}$ to achieve a predefined quality $Q$ as much as possible.

The camera locations and orientations $C = \{\{p_1, \varphi_1\}, \{p_2, \varphi_2\}, ..., \{p_N, \varphi_N\}\}$ constitute a 6D search space. Generally, roll angle could be neglected in practice, resulting in a 5D search space for camera poses. Some researchers also add some more constraints (e.g. restricting the search space to a sphere around the object (Vasquez-Gomez et al., 2009), some adaptive planes (Peng and Isler, 2019), or height-wise data capturing (Sharma et al., 2019)). They further reduce the computations at the cost of having less chance to find the optimal solution. Various heuristics $\mathcal{H}$ (cf. section 1.2 ) could be utilized to encourage the pose optimization algorithms to converge to a solution that meets the requirements of each specific application.

There are also various and different constraints $\Omega$ which could be considered during path planning. Battery life, the shortest path between viewpoints, detecting and avoiding permanent and moving obstacles, and excluding no-go areas in the scene are the main constraints that are investigated.

Predefined quality $Q$ of 3D reconstruction depends on the objective of the project and could be different in specific domains and applications. For example, in some applications wherein the full coverage of 3D reconstruction is very important, $Q$ could be formulated as the completeness of the results and other metrics like the accuracy of the reconstructed object could be relaxed. On the other hand, in some domains in which the accuracy is very important, other constraints have less priority, and accuracy has the highest weight in the planning. We will address these metrics in Section 3.

## 2 Viewpoint and path planning strategies

Given that a maneuverable platform (e.g., UAV in our scope) is available, an initial tempting idea could be to capture so many images to achieve the best 3D reconstruction. Considering the limited endurance of the UAVs and also other practical flight-time limitations, diminishing returns and even destructive effect of adding views after some points (Hepp, 2018; Hornung et al., 2008; Seitz et al., 2006) and high computation cost of processing all images, this method is neither efficient nor applicable in real scenarios. Hence, most UAV-based reconstruction projects are currently performed either using manual piloting or using off-the-shelf commercial flight planning applications with the camera pointing in a fixed direction (Smith et al., 2018; Yan et al., 2021). Another idea could be to select/filter a subset of already densely captured images for 3D reconstruction to decrease the significant computational overheads and avoid the destructive effect of some images on the final product. These techniques require a complete set of already captured images to filter some of them to achieve high quality 3D reconstruction. (Mauro et al., 2014), which employs view importance measure for filtering the viewpoints and (Maldonado et al., 2016), which is incremental NBV-based approach are two examples of this category. Besides data capturing time which still is a problem, the total computational cost of viewpoint selection and processing the remaining images could be close to the processing time of all images. Furthermore, camera locations are densely sampled and the other three parameters of camera pose namely rotations are not optimized. Some other approaches try to improve the output of regular pattern data capturing by estimating the missing parts and low-quality 3D reconstructed areas and iteratively adding some amending viewpoints to improve the final 3D product (Zhang et al., 2020). However, the best solution could be to have a holistic planning approach before starting data capturing which is the topic of a rich body of viewpoint and path planning literature in various research communities.

Automated flight planning methods can be classified either as model-free or model-based methods (Jing et al., 2016; Kaba et al., 2017; Koch, 2020; Yan et al., 2021; Zhang et al., 2020). Model-based methods use an initial coarse model (geometric proxy or simply proxy) for generating optimal viewpoint and path planning. But, model-free methods tackle the problem given that there is no such initial coarse model of the objects and environment and iteratively generate and update the model with new measurements. In the following parts of this section, we briefly discuss off-the-shelf mission planners. Afterward, we delve into the model-free and model-based methods.

### 2.1 Off-the-shelf flight planners:

The manual piloting is inevitable in some applications (Maboudi et al., 2021). However, the most common method for UAV-based data capturing in an automated fashion is to use an off-the-shelf flight planner, such as Pix4D capture[2], Altizure[3], Litchi[4], DJI Terra[5], DJI flight planner[6], 3dr robotics[7], Precisionhawk[8], PixHawk Ardu mission Planner[9], DroneDeploy[10], precision flight, UgCS[11].

Usually offering limited parameter settings, commercial flight planners generate conservative trajectories (e.g., an orbit, a zigzag, or a lawnmower pattern at a safe flight height) to cover the scene (Koch et al., 2019; Kuang et al., 2020; Peng and Isler, 2019; Roberts et al., 2017). However, these trajectories are generated regardless of the scene geometry and structure and distribution of the objects. Hence, these tools tend to over-sample some regions (e.g., rooftops) while under- sampling others (e.g., facades, convex areas, overhangs, and fine details), especially in dense areas and complex objects, and therefore sacrifice reconstruction quality (Hepp et al., 2018b; Roberts, 2019; Zhou et al., 2020b). Furthermore, most of such planning tools do not directly choose viewpoints adhering to all SfM and MVS constraints and do not directly account for complete coverage of the scene. Recently Agisoft introduced a tool for designing an optimum path-planning for image capturing and creating mission plans based on the rough model (Agisoft, 2022; Zhang et al., 2020). In this tool, a set of camera positions are generated to cover the object's surface with sufficient overlap. However, it cannot promise the completeness of the dataset in real-world situations, and the accuracy of the final point is not explicitly considered in this approach. Although the Agisoft mission planning tool has a power-line detection algorithm, most off-the-shelf flight mission planners may cause an accident with adjacent obstacles in the environment and should be used cautiously and at the safe fixed distance to the object.

---

[2] https://www.pix4d.com/product/pix4dcapture
[3] https://www.altizure.com/
[4][4] https://flylitchi.com/
[5] https://www.dji.com/de/dji-terra
[6] https://www.djiflightplanner.com/
[7] https://3drobotics.com/
[8] https://www.precisionhawk.com/
[9] https://ardupilot.org/planner/index.html
[10] https://www.dronedeploy.com/product/mobile/
[11] https://www.ugcs.com/

*2.2  Model-free methods*

Model-free methods do not have any prior information for target structures or scenes. Since there is no prior information, it is challenging to compute an optimal scanning trajectory. Therefore, model-free methods have to find the best scanning trajectory in an online manner from a partially constructed model. This problem is the same as the exploration planning problem, which determines the scanning paths online to explore an unknown and spatially bounded space. In this problem, a scanning platform has to estimate its location and construct 3D models in real-time. The location can be estimated using the simultaneous localization and mapping (SLAM) method (Leonard and Durrant-Whyte, 1991), and sparse 3D models can be constructed using an environmental mapping method (Estrada et al., 2005). SLAM systems are used to define the poses of the platforms, capture data from the scene, and direct the platform to the defined positions. Still, before directing the platform, the pose of the next station must be determined. In the last two decades, many SLAM approaches used a probabilistic method for reducing the impact of the inaccurate sensor on the map, using millimeter waves for building relative maps (Dissanayake et al., 2001), integrating Particle Filter and Extended Kalman Filter (Montemerlo et al., 2002), or introducing Square Root Smoothing and Mapping (Dellaert and Kaess, 2006). Another SLAM system is Visual SLAM which can produce a fast 3D reconstruction on-the-fly (Fang and Zhan, 2020).

Most model-free methods employ the next-best-view (NBV) approach. This approach is a greedy method that finds a local solution from partial information. It consistently determines the best viewpoint that obtains the largest unknown information from a current map. The unknown information is defined according to the purpose of various scanning scenarios. Some methods (Batinovic et al., 2021; Cieslewski et al., 2017) try to explore a volumetric map within a minimum time thoroughly. They evaluate the unknown volumes to determine an NBV in the volumetric map. Some methods (Bircher et al., 2018; Song et al., 2020a) are trying to inspect all surfaces of an unknown structure. These methods find a viewpoint that observes the largest uncovered surfaces. The other methods (Hardouin et al., 2020; Kompis et al., 2021; Song and Jo, 2018) try to reconstruct precise 3D models of target structures or environments. These methods analyze the reconstruction quality of 3D models, such as completeness and accuracy (Knapitsch et al., 2017). This section classified the model-free methods into three categories: *frontier-based planning*, *volumetric-based planning*, and *surface-based planning*. The following sections present literature reviews on each category of model-free planning methods summarized in Table 1.

2.2.1  Frontier-based methods

Frontier-based planning is one of the most widely used mobile robot exploration methods. The main idea was originally proposed by Yamauchi (Yamauchi, 1997), who defined the frontier as a boundary between an explored and unexplored area in a 2D occupancy grid map (Moravec and Elfes, 1985). The occupancy grid map represents a 2D environment as a set of grid cells composed of three states: free, occupied, and unknown. The frontier cells are estimated by collecting free cells adjacent to unknown cells. The frontier-based method starts at an initial location with an initially scanned map. The method continuously moves toward the nearest accessible and unvisited frontier cell until entire unknown cells are explored. This method generally shows acceptable exploration performances while intuitive and straightforward. Umari and Mukhopadhyay (2017) extended the frontier-based method to detect the frontiers in the occupancy grid map efficiently. They found frontiers by expanding multiple rapidly-exploring random trees (RRTs) instead of classical image processing methods like edge detection. The RRTs expanded towards unexplored regions based on the Voronoi diagram; they quickly extracted the frontiers with navigation paths.

The frontier-based method has been extended to 3D exploration tasks of a UAV by utilizing a 3D volumetric map (Hornung et al., 2013) instead of a 2D occupancy grid map. Frontier volumes are defined as the free volumes adjacent to the unknown volume in volumetric maps. The octree structures can rapidly access the 6- or 18- adjacent volumes. Most methods iteratively determine the NBVs based on the frontier information. Oßwald et al. (2016) improved the exploration performance of the general frontier-based method by providing additional guide information on global routes. They utilized a user-provided topological graph representing the topology information of the environment. A global route is computed by finding a traveling salesperson problem (TSP) solution on the topological graph. A mobile robot sequentially visits every node based on the TSP tour while exploring local regions using the frontier-based method. This method can produce more effective exploration paths that prevent revisiting a region that has already been passed. Cieslewski et al. (2017) extended the frontier-based method to maintain the maximum speed of a mobile robot. They determined an NBV by selecting target frontier volumes inside the current field of view (FoV). This method can prevent a mobile robot from slowing down due to heading rotation; therefore, the robot can explore unknown regions at high speed. This method reduces the exploration time while it may increase the trajectory length.

Several studies (Batinovic et al., 2021; Dai et al., 2020) have focused on reducing the computation time of frontier extraction and clustering. Dai et al. (2020) proposed an implicit frontier clustering method. They updated the frontiers only in the volumes inside the camera frustum at every depth integration step. Batinovic et al. (2021) tried to filter out many frontier points in multi-resolution volumes on Octomap. They also applied the mean-shift clustering for frontier clustering. These approaches significantly reduce the number of frontiers, which reduces the computation time of candidate viewpoint evaluation.

Some others (Meng et al., 2017; Shade and Newman, 2011) have tried to determine the next-best trajectories instead of the NBVs. They evaluated view sequences and found the most informative view paths. Shade and Newman (2011) proposed a 3D exploration method for stereo-camera-based scanning. They first extracted frontier volumes in a volumetric map and composed 3D vector fields toward the frontier volumes. They then computed the steepest descent path in the vector fields to get the next-best trajectory for 3D scanning. Meng et al. (2017) proposed a two-stage planning method for 3D exploration. The method first generates frontier sets by clustering adjacent frontiers and then computes the coverage path to cover all frontier sets. The coverage path can be computed by solving a fixed start open TSP. A mobile robot moves to the first edge of the coverage path and continuously recomputes the coverage path. Since this method considers global coverage instead of a single viewpoint, the number of revisits of the same area can be reduced.

### 2.2.2 Volumetric-based planning

Volumetric-based planning focuses on reconstructing a complete volumetric model instead of a fast exploration of unknown areas. Unlike frontier-based planning, volumetric-based planning analyzes not only the frontier but also the entire spatial information of the volumetric map. Estimated 3D information is sequentially integrated into an octree structure (Hornung et al., 2013). Each volume has one of three states (free, occupied, or unknown volume), updated based on the occupancy probability. The volumetric map is very suitable for view planning since it can easily access entire spatial information by the multi-resolution representation of Octomap. Furthermore, it can efficiently perform a ray-casting for visibility check. This operation is essential for viewpoint evaluation.

Volumetric-based planning method has also been widely used in object modeling. Vasquez-Gomez et al. (2017) proposed a volumetric-based planning method for single object modeling. They evaluated viewpoints by analyzing unknown volumes with overlapped volumes for point cloud registration. They also applied a hierarchical ray-tracing method for fast visibility check. This method starts from the rough-resolution map and applies high-resolution ray-tracing only around occupied volumes. Delmerico et al. (2018) proposed several information gain functions for viewpoint evaluation on the volumetric model. To quantify the visible information, they just counted unobserved voxels or calculated a weighting sum of voxel entropy. The authors also considered the object's rear side voxels with their entropy. They evaluated the performances of the proposed information gain functions through various modeling scenarios. Daudelin and Campbell (2017) proposed a view planning method for modeling an object without any prior information such as size or bounding box. They dynamically extended search spaces for the sampling of viewpoints based on a partial reconstruction. The search spaces are not restricted to known free space. They estimated all reachable configurations of a mobile robot's sensor and generated viewpoint samples from the reachable configurations. Similar to (Delmerico et al., 2018), the total amount of information gain for each candidate viewpoint is computed.

Several studies also proposed a volumetric-based planning method for exploration tasks. Bircher et al. (2016) employed a receding horizon planning strategy for exploration planning. The strategy iteratively computes an optimal exploration path and only executes the first step of the optimal path. This method generates a set of viewpoints by expanding an RRT and selects the best branch that provides the maximum information gain. The information gain is defined as the total volume of visible unknown cells penalized by the distance costs of nodes. The method then determines the NBV to the first node of the branch. This approach can efficiently find exploration paths with a low-computation complexity even in large-size environments. Papachristos et al. (2019b) extended the receding horizon planning method (Bircher et al., 2016) to additionally consider the localization uncertainty in exploration tasks. The proposed method comprises two planning stages: volumetric exploration planning and uncertainty-optimization planning. Like the receding horizon planning, the exploration planning expands branches of a random tree and determines an NBV from the first node of the most informative branch. To compute an information gain, they consider not only unknown volumes but also the occupancy probability of occupied volumes. The uncertainty-optimization planning then finds an optimal trajectory to the NBV, which provides the minimum localization uncertainty for the SLAM module. Dharmadhikari et al. (2020) also extended the receding horizon planning to provide fast and agile paths for a MAV. They directly generated view configurations and admissible paths based on motion primitives. They then determined a future-safe path that provides the maximum information gain and guarantees continuous fast motions. Batinovic et al. (2022) proposed an exploration planning method for LiDAR scanning data. They applied a recursive shadow-casting algorithm for fast information gain computation of large-scale point clouds. They also proposed a cuboid-based path evaluation method that estimates the information gain of each RRT edge instead of a node.

Song and Jo (2017) applied an inspection approach to a model-free planning method. Given a prior model of a target structure, the inspection approach precomputes a coverage path that provides full visual coverage of the whole surfaces of the prior model. On the other hand, Song and Jo (2017) addressed an online inspection planning on an incrementally updated and partially known model. The inspection planning method first determines an NBV that explores the largest unknown area in a volumetric map. Similar to (Bircher et al., 2016), it expands branches of RRT and evaluates the information gain of each branch to determine the NBV. The method then plans a local inspection path to the NBV, providing complete visual coverage of near-frontiers. The local inspection path is continuously replanned according to the updated near-frontiers until a robot reaches the NBV. This method improves the completeness of volumetric modeling because it can thoroughly scan all small unreconstructed regions. Song et al. (2020a) also extended the online inspection planning to consider global coverage planning. They proposed an online map partitioning method, which decomposes entire space into a set of sectors by clustering free and unknown volumes. The decomposed sectors have a compact and convex shape; therefore, the sectors represent a topological map. The method plans a global coverage path of unexplored sectors and determines an NBV to move toward the next sector. It then plans a local inspection path that fully covers local frontiers. This method significantly reduces total exploration time and path length by reducing the number of revisits of the sectors.

### 2.2.3 Surface-based planning

Surface-based planning concentrates on reconstructing a precise 3D surface model, which is represented as surface meshes (Kazhdan and Hoppe, 2013), surfel points (Whelan et al., 2016), or truncated signed distance function (TSDF) (Newcombe et al., 2011). Volumetric models are difficult to sufficiently express complex surface information; therefore, the surface-based planning methods analyze the shape and trend of reconstructed surfaces instead of volume information. Most surface-based planning methods have been used to reconstruct small-scale objects. Chen and Li (2005) predicted the shape and trend of simple and smooth surfaces from the curvature tendency and determined an NBV based on the predicted surface information. Wu Shihao et al. (2014) also predicted a tentative surface model from a partial reconstruction by using the Poisson surface reconstruction algorithm (Kazhdan and Hoppe, 2013). They then evaluated the completeness and smoothness of Poisson iso-surfaces to determine an NBV. Lee et al. (2020) detected surface shapes with surface primitives such as planes, cylinders, and spheres and utilized them for NBV determination.

Recently, several studies have attempted to apply the surface-based planning method to large-scale modeling tasks. Hardouin et al. (2020) analyzed the reconstructed surfaces to determine an NBV for MAV exploration. They estimated 3D data using a stereo camera or RGB-D sensor and integrated them into a surface model based on the TSDF (Newcombe et al., 2011). TSDF can reconstruct detailed 3D surfaces in real-time from a volumetric distance field, which contains sign distances to the closest surface. TSDF is also useful for identifying missing model parts during an online reconstruction (Monica and Aleotti, 2018). To determine an NBV, Hardouin et al. (2020) first detected incomplete surface elements (ISEs) from TSDF and then generated viewpoints along with normal directions of ISEs at a specified distance. They clustered the neighboring viewpoints and selected a cluster covering the most ISEs. They incrementally updated the TSDF model by scanning from an NBV until no ISE was detected.

Song and Jo (2018) proposed a surface-based exploration, which is an extended method of online inspection planning (Song and Jo, 2017) for surface model reconstruction. The surface-based exploration analyzes both the volumetric map and surface model in TSDF for fast exploration and complete reconstruction. The method determined an NBV that explores the largest unknown areas in the volumetric map. The method then plans an inspection path covering the surrounding low-confidence surfaces and frontiers. To cover the low-confidence surfaces, it generates a viewpoint set for each surface by inversely composing a view frustum from a surface to its normal direction. An inspection path is computed by finding the minimum distance trajectory that visits at least one viewpoint from each viewpoint set. A generalized TSP algorithm can compute the minimum distance trajectory. The method then refines the inspection path to cover the near-frontiers by applying the inspection planning method of Song and Jo (Song and Jo, 2017). Their method has a longer path length and more completion time, but the percentage of the coverage and the quality of the model are improved in their method. However, using a better objective function could improve the quality of the 3D reconstruction.

Schmid et al. (2020) proposed an exploration planning method for reconstructing a surface model. They implemented a new RRT method, which continuously expands a single tree to get global coverage with the maximum utility. Unlike the original RRT method, it rewires tree nodes according to their utility to maintain non-executed nodes and sub-trees. The method then refines intermediate paths and computes a global coverage path that maximizes the utility. They also proposed an information gain function for TSDF-based surface models. It considers not only a TSDF weight but also a sensing error. The sensing error of the depth sensor is modeled as the quadratic weight of the depth range. In their experiments, the proposed method can reconstruct more accurate 3D models than the other surface-based planning methods such as (Yoder and Scherer, 2016) and (Song and Jo, 2018).

A new informed sampling method for exploration path planning is introduced in (Kompis et al., 2021). The informed sampling method ranks the sampled viewpoints based on a utility function and finds the viewpoint candidates that are likely to have high information gain in advance. They used a stereo camera for 3D mapping the environment and Voxblox (Oleynikova et al., 2017) as the map representation of the scene. They generated viewpoints based on surface normal and frontier voxels' normal and computed their ranks according to an artificial potential field. The artificial potential field considers MAV position and viewpoint's distinction and repetitiveness. The method then evaluated only a subset of viewpoints according to the ranks. This approach significantly reduces the computation time of viewpoint evaluation.

Song et al. (2020b) applied a surface-based planning method to online multi-view stereo (MVS) reconstruction for the first time. On the other hand, Song et al. (2020b) implemented an online MVS system that reconstructs a large-scale scene in real-time. The system computes camera poses from the key-frame-based SLAM (Mur-Artal et al., 2015) and estimates depth maps of key-frames by performing the stereo matching of neighboring key-frames. For the depth estimation, it utilizes a monocular mapping algorithm, REMODE (Pizzoli et al., 2014). The estimated depth maps are integrated into a single 3D model based on the surfel mapping method (Whelan et al., 2016). They utilized the surface-based exploration method to plan a local inspection path for scanning low-confidence surfaces. Trajectory optimization is applied to maximize the MVS performance. They considered several multi-view stereo heuristics for the trajectory optimization, such as parallax, relative distance, and focus angle. Furthermore, Song et al., (2021a) extended this work for a more precise 3D model reconstruction. They utilized the deep-learning-based MVS, CasMVSNet (Gu et al., 2020), instead of REMODE (Pizzoli et al., 2014) for depth computation. CasMVSNet estimates a depth map by using multiple small cost volumes. It progressively reduces the depth hypothesis range in a coarse-to-fine manner, making it possible to process a high-resolution image in real-time. CasMVSNet provides better reconstruction performances with respect to accuracy and completeness than REMODE.

Surface-based planning analyzes detailed reconstructed surfaces in a surface model; however, the surface model may require a lot of memory. In this case, it can be limited to implementing the 3D scanning system onboard. To address this problem, a coarse surface model can be utilized for path planning by lowering the resolution of the reconstructed model. However, the advantage of surface-based planning that analyzes the surface shape and trend of a precise 3D model is lost. Finding a way to efficiently handle large-scale 3D data for an onboard system would be an interesting direction for future work. Table 1 Summarizes the state-of-the-art model-free approaches.

Table 1: Categorization and main features of model-free view-path planning methods

| Category | Approach | References | Modeling Method | Modeling Target | Description |
|---|---|---|---|---|---|
| Frontier-based planning | NBV planning | (Yamauchi, 1997), (Umari and Mukhopadhyay, 2017) | 2D occupancy grid mapping | Indoor/outdoor | Continuously move toward the nearest frontier in 2D grid map |
| | | (Zhu et al., 2018) | 2D occupancy grid mapping | Indoor | Deep reinforcement learning-based exploration planning |
| | | (Oßwald et al., 2016), (Cieslewski et al., 2017) | Volumetric mapping | Indoor/outdoor | Continuously move toward the nearest frontier in 3D space |
| | | (Dai et al., 2020), (Batinovic et al., 2021) | Volumetric mapping | Indoor/outdoor | Reduce the computation time of frontier extraction and clustering |
| | Next-best trajectory | (Shade and Newman, 2011), (Meng et al., 2017) | Volumetric mapping | Indoor/outdoor | Evaluate view sequences and determine the best trajectory |
| Volumetric-based planning | NBV planning | (Vasquez-Gomez et al., 2017), (Delmerico et al., 2018), (Daudelin and Campbell, 2017) | Volumetric mapping | Small-scale object | Provide various evaluation metrics for NBV planning on volumetric model |
| | | (Hepp et al., 2018a) | Volumetric mapping | Indoor/outdoor | 3D CNN-based viewpoint evaluation for NBV planning |
| | Next-best trajectory | (Charrow et al., 2015a), (Charrow et al., 2015b), (Wang et al., 2020) | Volumetric mapping | Indoor/outdoor | Evaluate or optimize view paths by using an information-theoretic measure |
| | Receding horizon planning | (Bircher et al., 2016), (Papachristos et al., 2019b), (Dharmadhikari et al., 2020), (Batinovic et al., 2022) | Volumetric mapping | Indoor/outdoor | Iteratively computes a next-best trajectory and only execute the first step of the trajectory |
| | Inspection planning | (Heng et al., 2015), (Song and Jo, 2017), (Song et al., 2020a) | Volumetric mapping | Indoor/outdoor | Online inspection path planning for complete modeling |
| Surface-based planning | NBV planning | (Chen and Li, 2005), (Wu et al., 2014), (Lee et al., 2020) | Surface reconstruction | Small-scale object | Surface-based NBV planning for modeling a small-scale object |
| | | (Hardouin et al., 2020), (Kompis et al., 2021) | TSDF mapping | Indoor/outdoor | Evaluate reconstruction quality for NBV planning |
| | | (Peralta et al., 2020), (Zeng et al., 2020) | Point cloud mapping | Single object | Machine-learning-based NBV planning for point-cloud mapping |
| | Receding horizon planning | (Schmid et al., 2020) | TSDF mapping | Indoor/outdoor | Iteratively rewire nodes of RRT for coverage planning |
| | Inspection planning | (Song and Jo, 2018) | TSDF mapping | Outdoor | Online inspection planning for dense surface reconstruction |
| | | (Song et al., 2020b), (Song et al., 2021) | MVS | Outdoor | View path planning via online multi-view stereo reconstruction |

*2.3 Model-based methods*

Model-based methods leverage a coarse representation of the scene geometry (geometric proxy) for generating optimal viewpoint and path planning. Recently, Agisoft Metashape 1.5 introduced functionality for designing an optimum path-planning for image capturing, which creates mission plans based on the rough model (Agisoft, 2022; Zhang et al., 2020). In this approach, a set of viewpoints with sufficient overlap is generated to cover the surface of the object. However, the computed static network of camera poses cannot promise the completeness of the dataset in real-world situations, and the accuracy of the final point cloud is not considered explicitly. Recently explore-then-exploit methods (Figure 2) which consist of two phases are utilized in UAV mission planning (Peng and Isler, 2019; Roberts et al., 2017; Yan et al., 2021; Zhang et al., 2020). The first phase is the *exploration* that generates the initial viewpoints and flight path similar to off-the-shelf flight planners or manual flight (Koch et al., 2019). The second phase is called the *exploitation* phase, where the geometric proxy is employed to design optimum viewpoints and trajectories to acquire proper images for 3D reconstruction (Sadeghi et al., 2019; Zhou et al., 2020b).

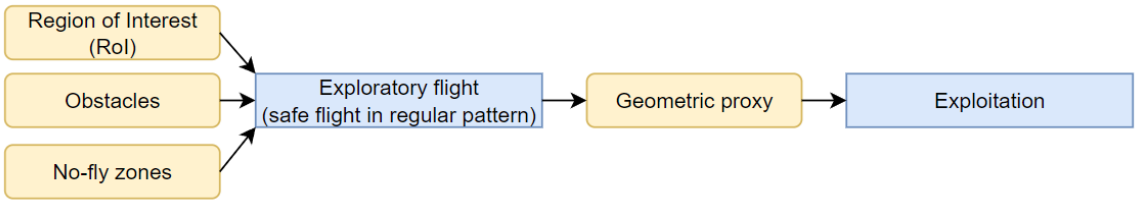

Figure 2: Overall structure of explore-then-exploit methods

The main advantage of the explore-then-exploit strategy is that prior knowledge of the scene geometry facilitates the global optimization of 3D coverage and accuracy in the secondary exploitation visit (Peng and Isler, 2019).

2.3.1 Exploration

Although most researchers use a nadir flight for this phase, employing orbit patterns is also reported in some literature (Roberts et al., 2017). The outcome of exploration phase approximates the scene geometry which is called "geometric proxy". The geometric proxy is usually a mesh (Hoppe et al., 2012; Sharma et al., 2019), point cloud (Yan et al., 2021), or supported by voxel representation (Hepp et al., 2018b). The initial model or geometric proxy obtained by exploration, may include many gaps or inaccurate areas (Hepp et al., 2018b; Koch et al., 2019). Therefore, this rough 3D representation of the scene is mostly used for optimal viewpoint planning (to generate a high-fidelity reconstruction) as well as computing the 3D safe navigable zone for the platform (Hepp et al., 2018b; Roberts et al., 2017; Smith et al., 2018). However, (Yan et al., 2021; Zhang et al., 2020) used this geometric proxy to locate incomplete or low-quality areas to guide the camera placement in the exploration phase.

Different from most other approaches, which use an exploratory flight for generating geometric proxy, in (Zhou et al., 2020a), a very coarse 2.5D coarse model is generated from the Google map (providing 2D footprints of buildings) and a single satellite image (for height estimation based on shadows).

Using this approach, the planning can be done before visiting the target sites. However, it needs an up-to-date satellite image captured during a sunny day and assumes that the ground is relatively flat. Moreover, extrusion of the 2D footprint of irregularly shaped architectural objects leads to a non-optimal viewpoint planning in the exploration phase.

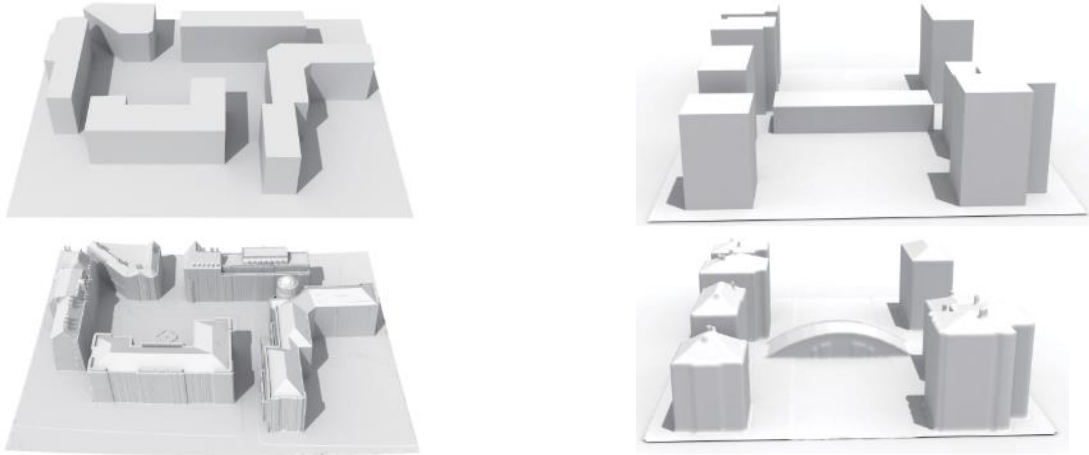

Figure 3: comparison of proxies. First row: coarse 2.D proxies generated in (Zhou et al., 2020a); Second row: 3D proxies generated using an exploratory flight. Images are adopted from (Zhou et al., 2020a)

A semantic-aware exploit-and-explore flight planning approach is introduced in (Koch et al., 2019). The main contribution of this research lies at employing semantic information of the exploratory flight for extracting the target object and generating a semantically-enriched coarse proxy for defining inadmissible airspaces for safe trajectories. A fully Convolutional Network (Long et al., 2015) is applied on 3069 images (from Semantic Drone Dataset[12] and ISPRS 2D Semantic Labelling Benchmark (Rottensteiner et al., 2014) and also some manually annotated UAV images). The open street map (OSM) information are used to improve the results. The object of interest is extracted using a region growing seeded by the user input, which is applied on the segmentation result. The semantics of the environment is very important (both practically and legally) in defining the flight trajectory, especially in dense urban areas.

It is worth mentioning that with the recent advancements in the concepts and realization of BIM, digital twin, and smart cities, the 3D rough models of large structures like buildings and bridges are going to be more available. Although this 3D spatial information can alleviate the exploration in UAV mission planning, semantic and geometric information of the whole scene should be updated for a safe UAV mission and high-fidelity 3D reconstruction.

### 2.3.2 Exploitation

After generating the geometric proxy, exploitation aims at designing optimum viewpoints and trajectories to provide the data for high-quality 3D reconstruction (Sadeghi et al., 2019; Zhou et al., 2020b) constrained by platform and environment conditions.

(Hoppe et al., 2012) proposed an uncertainty-aware and heuristic-based viewpoint planning approach which aims at accurate 3D coverage of the scene. Their approach belongs to the generate-and-test approaches which generates potential camera poses by assigning one fronto-parallel view to each triangle of the proxy-mesh. Then an angular histogram (four bins in [0-40]) is computed for each triangle. Another histogram which is the accumulation of cameras of all triangles for each angular bin is used to score the cameras. Based on camera scores (which are updated after the selection of each viewpoint), a greedy selection of the best (i.e. observes the most triangles at a novel angle) overlapping neighboring viewpoints is employed in order to select a subset of the viewpoints to reach one of the two stopping conditions for each (in practice 95% ) of the triangles: 1) estimated accuracy, measured by the largest semi-axis of the covariance ellipsoid, is acceptable, or 2) at least one camera of each bin is selected. Although the efficiency of this approach is not reported, we believe that initializing the camera positions by considering one candidate for each triangle of the proxy mesh may generate a very dense search space and is not applicable in real-world large-scale projects. Since this is an off-line approach, no feedback about texture or illumination conditions is considered during data capturing.

Jing et al. (2016) subdivided the 3D rough model using the Bubble Mesh method. The initial model is divided into small patches. Afterward, randomly sampled viewpoints are generated around the target building. After investigating the visibility (between each surface patch and each viewpoint) and adjacency (between each pair of viewpoints), the selection of viewpoints is performed as a modified set covering with constraints using a neighborhood greedy search algorithm. SFM and MVS heuristics are not involved in the viewpoint selection and computing the orientation of the cameras and quantitative evaluation of the results are missing. Although the proposed method for designing image orientations seems feasible, generating random positions is too risky, and it might cause some gaps or low-quality areas on complex objects.

---

[12] https://www.tugraz.at/index.php?id=22387

In the seminal research of (Roberts et al., 2017), selecting the viewpoint and trajectory planning are jointly considered in a unified global optimization problem as a submodular orienteering problem. This innovative approach finds a path that maximizes scene coverage while adhering to the SFM and MVS heuristics and respecting the limited travel-budget of the drone. After introducing a coverage model, the approximated optimal orientations of all candidate cameras are computed. Submodular orienteering is then transformed into an additive orienteering problem which is solved as an integer linear problem. However, the coverage model of this approach is monotone which means selecting more cameras never reduces the coverage score, which is not in line with the findings of (Hepp, 2018; Seitz et al., 2006). Moreover, there is no stopping condition except travel-budget, which is independent of the complexity of the object and desired quality. Lacking online processing also prevents reaction to low-texture areas and moving objects during flight. Lastly, uniform sampling of the object points avoids putting more attention on the edges.

Hepp et al., (2018b) proposed a system for 3D reconstruction of building-scale scenes which is similar to Roberts2017 in spirit. Based on the coarse mesh (generated from images with overhead pattern), a volumetric occupancy map (fixed voxel size=0.2m) containing occupied, free-space, and unobserved voxels is generated where each voxel is also attributed by a measure of observation quality. The authors introduced an approximated camera model (to make the optimization problem suitable for a submodular formulation) to measure expected information gain from an individual viewpoint, independently from other viewpoints. After generating viewpoints candidates, ray-casting is used to compute the visible voxels for each viewpoint to evaluate the corresponding contributed information. Since this model entirely ignores the MVS heuristics, the proposed method encourages fronto-parallel views during the computation of information contribution of voxels. To solve the submodular optimization with travel-budget constraint, an adaption of the recursive strategy which is introduced in (Chekuri and Pál, 2005) is used. This strategy recursively splits the trajectory and travel-budget into two parts and selects the reachable (with the current travel budget) viewpoint with the highest information gain. The authors employed voxels' attributes (free, occupied, unknown) to find the free-space motion paths. Paths between viewpoints are preferably a straight line, or in case of an obstacle they are piecewise linear computed using RRT* algorithm (Karaman and Frazzoli, 2011). In spite of very promising results, this approach uses images from the exploratory flight in reconstruction and also in uncertainty computation. This would lead to a problem when the geometric proxy is not generated from the exploratory flight (cf. section 2.3.1). Even if a good geometric proxy is available, the exploration should start after an initial flight. Another issue of these methods is that their optimization needs the minimum number of viewpoints as input and is not guaranteed to converge. The user should decide how long to run the optimization. Furthermore, using a recursive greedy algorithm to maximize the optimization objective may lead to some viewpoints which cannot make a single adjustable block. Hence, some other images are added to make sure that all the images are connected. (Hoppe et al., 2012; Mostegel et al., 2016) solve this problem by applying an overlapping constraint in the recursive selection process.

An Explore-then-exploit approach is introduced in (Peng and Isler, 2019) which after exploration and generating the geometric proxy, builds a set of adaptive viewing planes for viewpoint selection. This is an iterative approach in the exploration phase (usually one exploratory flight and two exploitation flights) which repeats data capturing after identifying the low-quality regions until quality converges or a desired level of quality is achieved. The main novelty of this approach which is an extension of their previous study (Peng and Isler, 2018), is considering a set of adaptive viewing planes (Figure 4) to limit the search space for camera placement leading to similar resolution for the images in each patch. However, generating viewing planes based on the slope of regions may cause gaps because viewing planes are created based on the average of patch points normal. Hence, in the complex regions the viewing plane might not be able to cover all regions, properly.

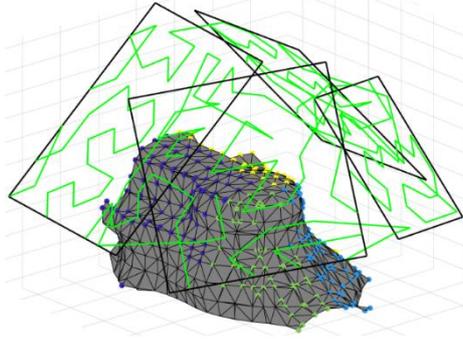

Figure 4: The adaptive viewing rectangles for different clusters of the scene patch (Peng and Isler, 2019)

In (Smith et al., 2018) a dilation operator with a ball structure element (whose radius is the desired minimum distance) is applied on the geometric proxy to estimate a closed surface boundary between safe and potentially dangerous airspace. The authors employed some reconstructability heuristics for predicting reconstruction quality. The contradicting effect of parallax on triangulation error and matchability are estimated and considered. Moreover, the distance of the camera to the object and binary visibility of the points based on the geometric scene proxy and the deviation from normal of the surface are also elaborated in the heuristic function. The authors use the geometric proxy to initialize the camera network with a fixed number of uniformly distributed cameras. Utilizing a set of uniformly distributed samples on the geometric proxy, pairwise view parameters are improved incrementally using the simplex method to optimize the reconstructability of the samples. Different aspects of 3D reconstruction heuristics are formulated and embedded very well in the objective function of this approach. However, since fixed and uniformly distributed sample points are used, depth discontinuities posed some problems in the experiments. Moreover, the inability of this off-line approach to adapt the number of views could be improved to avoid oversampling and undersampling of the viewpoints. Furthermore, like almost all other approaches, varying surface texture and lighting are not considered in their approach.

Sharma et al. (2019) introduced a floor-wise viewpoint and path planning for optimal building surface coverage. The geometric proxy is a mesh which could be obtained either from building blueprints or cross-sections of a rough model generated by a reconnaissance (exploratory) flight. For each height a contour is generated and minimum 60% overlap between image strips is considered to cover the contours in 3D. Offsetting the contours by a fixed distance, a 2D flight map is generated. This initial flight path could be updated to avoid obstacles which are computed with MVS reconstruction from the reconnaissance flight. The camera positions are initialized by sampling the flight path every 1 m and camera poses (position and orientations) are then optimized using genetic algorithm to maximize the total coverage such that at least N (=3) cameras see each patch of the surfaces. Since this approach does not consider most MVS heuristics, the generated 3D model is suitable for the applications like VR which the details and completeness of the dataset are more important than the accuracy of 3D reconstruction.

In (Yan et al., 2021), by evaluating the quality of the geometric proxy (point cloud), two types of new viewpoints are selected in a constrained view sampling space. The new viewpoints of first type are considered for improving incomplete or low-quality areas and the second type viewpoints are added to fully cover the entire scene. Finally, various optimization algorithms are employed to generate a smooth 3D trajectory. For this step, ACO, PSO, and Branch and Bound (BnB) methods are compared and the best results are reported using ACO. Similar to (Smith et al., 2018) the sample points on the geometric proxy are distributed uniformly. Therefore, most sample points with lower quality scores were located at the edges of buildings or structures with more geometric details. Two main objectives in their method are maximizing the coverage and minimizing the path length, so the quality of the 3D reconstruction does not play any role in their method for viewpoint planning.

(Kuang et al., 2020) proposed an on-line path planning method for large-scale urban scene reconstruction. In this research the geometric proxy of the buildings is generated using an oblique initial flight around the buildings. Then, instances in the scene are segmented using the well-known Mask-RCNN (He et al., 2017) approach and simple geometric shapes are used to represent the layout of the scene. Flight trajectory is minimized in two levels, namely inter-buildings and intra-building levels. In the first level of optimization each building is considered as a node of a graph (Figure 5) and the shortest path is computed using the classical Dijkstra algorithm. In the second level, the task is formulated as the Chinese postman problem.

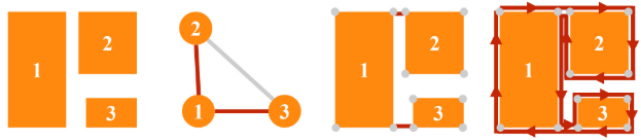
Figure 5: Bi-level graph-based path planning in (Kuang et al., 2020)

For on-the-fly predicting the coverage of the whole scene a lightweight SLAM framework is utilized. Then, unobserved spaces are covered by adding new viewpoints. If the endurance of UAV allows, this online approach also has a mechanism which tries to capture close-up images of architectural details. The main limitation of this approach is that SFM and MSV heuristics are not considered in the planning, directly. Moreover, since prior information of the coarse scene model is not completely used during flight, more viewpoints are required.

(Huang et al., 2018) proposed an efficient viewpoint and path planning method based on a fast MVS and NBV algorithm. This approach consists of an online front-end for viewpoint and path planning to control the image capturing process and offline back-end for generating the final dense 3D model (Figure 6). The online front-end incorporates a closed-loop for iterative updating the initial model, coverage evaluation and NBV planning to achieve satisfactory coverage.

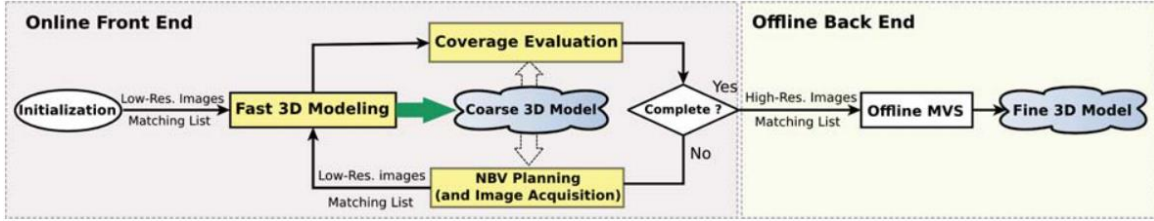
Figure 6: The pipeline of the proposed approach in (Huang et al., 2018). Image courtesy: (Huang et al., 2018)

The proposed approach uses the initial model generated from an enclosing rectangular flight around the object. A confidence score is computed for each triangle of the mesh. In this confidence score position consistency, normal consistency, and front parallelism are considered. Searching for NBVs is done plane by plane, from high to low altitude to speed up the process by coarsening the search space from 5D to 4D. In each iteration, an obstacle-aware A* algorithm is used for optimal path planning, which connects the NBVs starting from UAV's current position. The main novelty of this approach is using on-the-fly incremental SFM for local and per-vertex updating the mesh model. However, it highly relies on GNSS signal availability which is not always the case, especially in dense urban areas and close to the large objects.

Koch et al. (2019) proposed a semantic-aware exploit-and-explore flight planning approach to define inadmissible airspaces for safe trajectories. The main contribution of this research lies in employing semantic information of the exploratory flight for extracting the target object and generating a semantically-enriched coarse proxy for defining free and occupied airspace and avoiding prohibited flight zones. After sampling a large number of viewpoints on the bounding box of the extracted object of interest, the normalized distance-weighted mean of all visible 3D points is utilized to compute the orientation of each viewpoint candidate. Distance-based and observation angle-based gains coupled with camera constellation hemisphere slices on each object point are used as heuristics to predict the suitability of the viewpoints for the reconstruction. Inspired by (Hepp et al., 2018b; Roberts et al., 2017), greedy submodularity optimization is utilized to add the viewpoint with the highest additive information reward to the output set.

Table 2 summarizes the main characteristics of discussed model-based approach.

Table 2: Main characteristics of discussed model-based approach.

| | Exploration | Exploitation | | Exploitation | Path planning | On-line processing |
|---|---|---|---|---|---|---|
| | | pose candidate generation | | pose computation | | |
| | | position | angles | | | |
| (Hoppe et al., 2012) | Nadir flight | Centers of mesh triangles | Fronto-parallel to the mesh normals | Cluster then select | Greedy shortest path | No |
| (Huang et al., 2018) | Flight on an enclosing rectangle | Plane-wise 2D grid | Regular sampling of pitch and yaw | NBV based on confidence score (function of position consistency, normal consistency, and front parallelism.) | A* | Yes |
| (Jing et al., 2016) | 2D map and estimated building height | Random | Surface patch attraction forces | coverage ratio and fully connected viewpoints constraints | No | No |
| (Peng and Isler, 2019) | Nadir flight | viewing grid on each adaptive viewing plane | Normals of the adaptive viewing planes | SfM and MVS heuristics | patch-wise TSP | No |
| (Roberts et al., 2017) | Orbit flight | uniform sampling on 3D bounding box | Constrained uniform sampling | Based on the optimum path length | | No |
| (Smith et al., 2018) | Nadir flight | Uniformly distributed views | Randomly oriented simplices around each view candidate | Simplex optimization spherical visibility histograms | TSP | No |
| (Yan et al., 2021) | Nadir flight | centers of view sampling voxels | Normal angle of sample points | Local (confidence score (visibility, distance, sample points normal)) and Global (coverage) | TSP-ACO, PSO, BnB | No |
| (Kuang et al., 2020) | Oblique images | Looking at center of unobserved cells cluster | encourages fronto-parallel views | sorted confidence score (function of distance, number of views, and front parallelism) | Dijkstra | Yes |
| (Hepp et al., 2018b) | Nadir flight | regular 3D sampling Add viewpoint s to exploration queue | Random sampling with bias towards the interesting area | Information Gain score and Sparse matching | RRT* | No |
| (Zhou et al., 2020a) | GoogleMap+satellite image | one candidate for each sampling point | Fronto-parallel to the sampling poin | MaxMin optimization: minimizing the redundancy and maximizing the reconstructability | TSP | No |
| (Koch et al., 2019) | Nadir flight | uniform sampling on 3D bounding box | Normalized distance-weighted mean of 3D points | Submodularity (distance-based and observation angle-based reward) | orienteering | No |
| (Sharma et al., 2019) | Nadir flight | uniform sampling on height-wise boundaries | encourages fronto-parallel views | Genetic algorithm | Fast marching | No |

Since a prior model of the object is available, model-based methods are able to compute the reasonable poses that could be used to create a complete and accurate 3D reconstruction. Employing various objective functions which include SfM and MVS heuristics, accuracy and completeness are involved explicitly in selecting optimum viewpoints. Most of these approaches use monocular images and tackle the problem of viewpoint and path planning separately i.e. after computing the effective viewpoints, path optimization will start to compute an efficient solution to approach the computed poses. As Table 2 shows, most model-based methods are offline. An interesting direction could be using a fast online SfM and MVS system to enable on-the-fly computations.

To the best of our knowledge, all model-based approaches use uniform sampling of the proxy for further inspections or use a uniform discretization of the environment as 3D representation of the object of interest. Investigation of the results of the current approaches reveals that most of the large errors lie at highly curved parts and also sharp corners of the objects. We suggest considering an adaptive sampling of the objects where the overall density depends on the desired LoD. However, the density should be also adaptive to local curvature which leads to more samples on the high curvature areas and geometric boundaries.

The Static environment is the prerequisite of the most MVS algorithms (Mostegel et al., 2016). Therefore, detecting and considering moving objects during data capturing is very important. However, this issue is widely ignored by most planning and image acquisition approaches. Another limitation is low-texture areas and building parts with different materials. In some cases, especially for buildings, there might be some low-texture images that can't reconstruct the 3D model of the area or object. None of the above-mentioned model-based methods consider neither moving objects nor low-texture images for viewpoints planning.

Although some researchers have already employed Google Map and satellite images for generating geometric proxy, with the recent advancements in the concepts and realization of BIM, digital twin and smart cities, the 3D rough model of the large structures like buildings and bridges are going to be more available. Even LoD2 information is publicly available in some countries nowadays[1,2] that could be used as *a-priori* information in viewpoint and path planning algorithms.

## 3 Evaluation strategies

Evaluation of viewpoint and path planning methods for UAV-based 3D reconstruction is a challenging task which poses many significant challenges (Debus and Rodehorst, 2021; Hepp et al., 2018b; Koch et al., 2019). First, the important criteria in various communities and projects are different. For example, while visual fidelity and completeness are important in some cases, accuracy is a key in some other applications. Moreover, in some projects and environments, overall flight time is critical. Second, environmental conditions such as lighting, weather conditions, and surrounding objects may change. Moreover, moving objects and modifications of the object itself that can occur over the course of time can affect the evaluations. Furthermore, ground truth information is typically not available (Debus and Rodehorst, 2021; Hepp et al., 2018b; Koch et al., 2019). Hence, in most studies, the evaluation of the real datasets and scenes are restricted to the qualitative verification. Quantitative evaluations are usually possible and performed on synthetic datasets (cf. section 3.2) or specific dataset is used which make the fair and standard comparison of different approaches almost impossible. Last but not least, in most evaluations the quality of the generated 3D data (e.g. point cloud) is investigated. This indicates the performance of the whole pipeline of planning, data capturing and 3D reconstruction (Figure 7), which is the non-separable mixture of these three steps. The interested reader can refer to (Seitz et al., 2006) and the leader-board of (Knapitsch et al., 2017) for a comparison of the reconstruction approaches. It should be accentuated that even in the case of perfect planning, the effect of hardware and environment could affect the quality of the acquired images.

---

[1] https://3d.bk.tudelft.nl/opendata/opencities/
[2] https://opengeodata.lgln.niedersachsen.de/#lod2

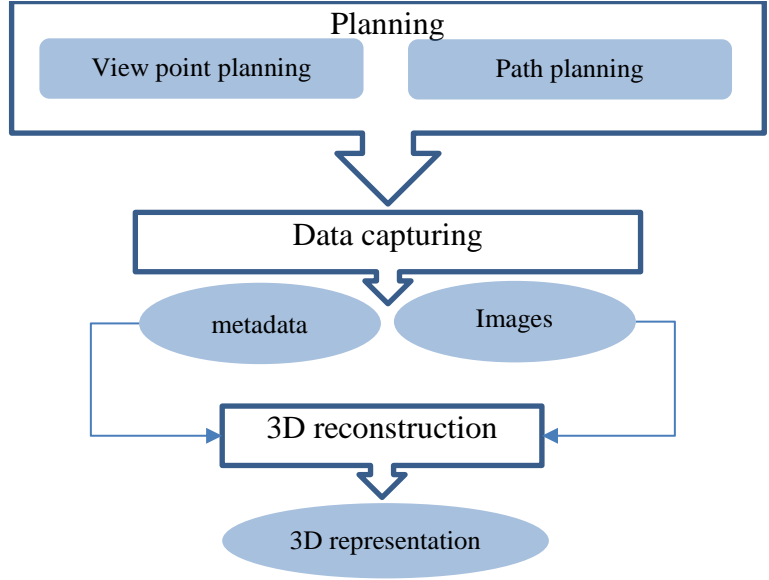

Figure 7: General pipeline of planning, data capturing and 3D reconstruction

Therefore, in this section we first mention the quality measures for the whole pipeline and then investigate measures that indicate the quality of viewpoint and path planning separately.

*3.1 Evaluation of the whole pipeline:*

Denoting the ground truth as $G$ and reconstruction result as $R$, we can compute the following quality measures:

- *precision* (how much of $R$ is supported by $G$)
- *recall* (how much of $G$ is correctly modeled by $R$)
- *F-score* (Harmonic mean of *recall* and *precision*)
- *accuracy* (how close $R$ is to $G$)

Generally, both $R$ and $G$ could be a mesh or point cloud. However, considering $R$ as point cloud is better (avoiding surface reconstruction and its errors) and more convenient (avoiding sampling mesh at its vertices (Seitz et al., 2006)). If $G$ is delivered as a mesh or CAD model, it is also better to use it directly to avoid the approximations in resampling it to a point cloud. Nevertheless, if $G$ is itself a point cloud (e.g. captured by laser scanners) we can directly use it in the point-based evaluations. Following the similar notations as in (Knapitsch et al., 2017) and (Seitz et al., 2006), the quality measures could be expressed as follows:

**Precision** (Correctness): if the distance between a reconstructed point $r \in R$ and the ground truth is $e_{r \to G} = \min_{g \in G}\|r - g\|$, the *precision* of the reconstruction for any cut-out distance $d$ is defined as follows:

$$P(d) = \frac{\sum_{r \in R}[e_{r \to G} \leq d]}{|R|} \qquad (1)$$

where $[\cdot] \in \{0,1\}$ is the Iverson bracket and $|\cdot|$ is the cardinality of the set and $P(d) \in [0,1]$. Selecting the proper value for $d$ depends on the application and affects the results. For example, (Hepp et al., 2018b) considered $d = 10\ cm$ which may not be acceptable for some high quality reconstruction purposes.

**Recall** (Completeness): Similarly, for a ground-truth point $g \in G$, its distance to $R$ is defined as $e_{g \to R} = \min_{r \in R}\|g - r\|$ and the *recall* of the reconstruction $R$ for any cut-out distance $d$ is $R(d) \in [0,1]$ and is defined as:

$$R(d) = \frac{\sum_{g \in G}[e_{g \to R} \leq d]}{|G|} \quad (2)$$

***F-score***: summarizes the *precision* and *recall* in a single score as:

$$F(d) = \frac{2P(d)R(d)}{P(d) + R(d)} \quad (3)$$

$F(d) \in [0,1]$ has a very important property that if either $P(d) \to 0$ or $R(d) \to 0$ then $F(d) \to 0$, which is not always valid for arithmetic mean.

*Accuracy:* Various metrics using unsigned and signed histograms of distances from $R$ to $G$ could be used to measure the accuracy of the reconstruction. While mean and standard deviation of distances could be used (Huang et al., 2018), the median of the distances is more robust to outliers. Another statistic which is suggested in the literature (Seitz et al., 2006) is the distance $D$ such that $X\%$ of the points on R are within distance $D$ of G. Median is a special case of this measure when $X = 50$. In (Seitz et al., 2006) $X = 90$ is used for computing the accuracy.

The main benefit of using signed distances compared to unsigned distances is that it gives a better sense of where and in which directions the reconstructed surface deviates from the ground truth (Maboudi et al., 2018; Seitz et al., 2006). To this end, one can visualize the distribution and inspect the trend of the signed deviations. Moreover, using unsigned distances changes the distribution of errors. In the case of normally distributed signed errors, the unsigned errors follow the folded normal distribution (Tsagris et al., 2014), whose parameters are different from the normal distribution's parameters.

To the best of our knowledge, there is no established framework for the evaluation of viewpoint planning, directly. The most common way is to report one of the metrics mentioned in the previous section as a function of the number of images for the 3D reconstruction of the object (Huang et al., 2018; Kuang et al., 2020; Peng and Isler, 2019; Smith et al., 2018). However, since these metrics are a function of both viewpoint planning quality and 3D reconstruction quality, these strategies cannot evaluate the viewpoint planning separately. The topic of establishing an evaluation pipeline and proper quality criteria is discussed in (Debus and Rodehorst, 2021). Based on the proposed pipeline, the authors evaluated some 3D UAS flight path planning algorithms. This pipeline focuses on pre-planned flight paths that have been computed based on available rough 3D models and is not transferable to planning in unseen and unknown environments that need to be explored.

Path planning quality which aims at minimizing the mission duration is also highly correlated with the number and distribution of viewpoints. Therefore, reporting mission duration as a function of measures like recall and precision (Kuang et al., 2020) depends also on view planning strategy. However, using the same viewpoints, path planning quality could be evaluated by measuring the flight path length and data capturing time. UAV battery duration and the speed at which the images can be captured are the main constraints that should be considered in path planning (Smith et al., 2018). We summarized the evaluation measures reported in the literature in Table 3. For the sake of brevity, just model-based approaches are listed.

Table 3: Evaluation measures reported in the model-based studies

|  | **Geometric measure** | **Efficiency measure** |
|---|---|---|
| (Hoppe et al., 2012) | Precision: C2M[1] (on synthetic dataset) | Number of images |
| (Huang et al., 2018) | Recall: C2C[2]<br>Accuracy: C2C (mean and RMSE) | Flight time, and<br>number of images |
| (Peng and Isler, 2019) | Recall: based on depth images<br>Accuracy: Mean and standard deviation of the depth | Number of images |
| (Roberts et al., 2017) | Recall: C2C<br>Accuracy: C2C | Travel budget (fixed) |
| (Smith et al., 2018) | Recall: C2M (on synthetic dataset)<br>Accuracy: C2M (on synthetic dataset) | Flight time, and<br>Flight path length |
| (Yan et al., 2021) | Recall: Voxel occupancy<br>Accuracy: C2M (on synthetic dataset) | Flight time, and<br>Flight path length |
| (Kuang et al., 2020) | Recall: C2M (on synthetic dataset)<br>Accuracy: C2M (RMSE) | Flight time, and<br>Number of images |
| (Hepp et al., 2018b) | Precision: C2C<br>Recall: C2C | Flight path length and<br>Number of images |
| (Zhou et al., 2020a) | Recall: C2M<br>Accuracy: C2M | Flight time, path length<br>and Number of images |
| (Koch et al., 2019) | Precision: C2C<br>Recall: C2C | Flight path length and<br>Number of images |
| (Sharma et al., 2019) | Accuracy: Hausdorff distance (mean and RMS) | Flight time |

There are also some attempts to combine different quality metrics to provide a unified score for both viewpoint and path planning steps to make different approaches comparable. In (Debus and Rodehorst, 2021) an evaluation pipeline is introduced which employs a normalized score for weighted combination of viewpoint and path planning quality as:

$$\hat{S} = \Psi \frac{A}{L.d^2} \qquad (4)$$

where $\Psi$ is the quality factor of the flight path, $L$ is the cost of the flight path length, $A$ is the surface area of the 3D model, and $d$ is the distance between camera and object. However, there is no optimal value for this normalized score and it could be used just for a relative comparison of the planning approaches. Moreover, some hyper-parameters should be added to equation (4) to tune the importance of viewpoint and path planning quality measures for different applications.

*3.2 Outdoor 3D reconstruction benchmarks*

Preparation of reliable ground truth for the whole scene is not a straight-forward task. Usually, more than one sensor should be utilized (Bouziani et al., 2021) to capture the entire object with the accuracy of at least one level of magnitude better than the data under investigation. Therefore, many researchers perform the qualitative evaluation on real datasets, while quantitative evaluation is performed on synthetic datasets (Hepp et al., 2018b; Yan et al., 2021).

---
[1] Cloud to Mesh

[2] Cloud to Cloud

Unreal Engine (UE[1]) is a game development engine that produces photorealistic scenes. UE is by far the most common simulation environment which is used in many researches in viewpoint and path planning field (Hepp et al., 2018b; Kuang et al., 2020; Peng and Isler, 2018; Roberts et al., 2017). An open-source Python library called UnrealCV (Qiu et al., 2017) could also be used to set the camera parameters and control a virtual camera in the simulated environments (Peng and Isler, 2019; Roberts et al., 2017). Since there is full control over the camera parameters, objects (like buildings) and also lighting and texture it is possible to create many different scenes and reconfigure their attributes that is a very useful tool for evaluation purposes. Smith et al. (2018) created a benchmark which consists of 34 uniquely different buildings in five large urban scenes and packaged them within a release project of UE, which is publicly accessible via their project website[2]. The authors developed a new synthetic benchmark dataset and simulation environment for UAV path planning problem in urban environments making it possible to quantitatively evaluate other viewpoints and path planning algorithms. A small part of one of synthetic samples from this benchmark is illustrated in Figure 8a. Microsoft AirSim (Shah et al., 2017), an open-source UAV simulator built on Unreal Engine, is also employed in the literature to simulate the flight process and plan the UAV flight path (Arce Munoz, 2020; Kuang et al., 2020).

Unity game engine[3] is another environment that is employed in (Sharma et al., 2019) to create a 3D scene (see Figure 8b). Scripting capabilities of the game engine provides the possibility to capture the images using a virtual camera. An experimental Unity release of AirSim is also announced[4]. In some studies like (Koch et al., 2019) the authors generated their synthetic dataset (see Figure 8c.) using other environments e.g. open-source computer graphics software Blender (Blender_Online_Community, 2018) which is a well-known 3D modeling and rendering package. 3D architecture models from the 3D Ware-house of SketchUp[5] are also used (Huang et al., 2018). The authors used the Gazebo robot simulator[6] for simulation and controlling the platform.

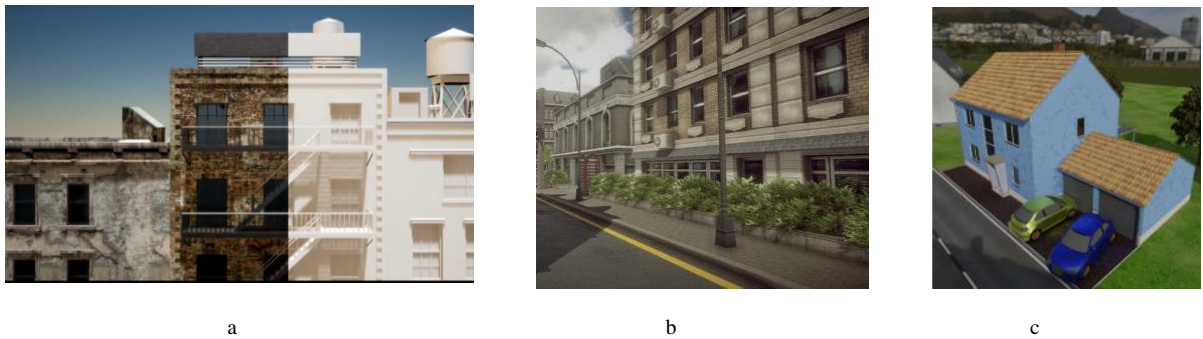

a          b          c

Figure 8: Three examples of synthetic environments: a) Part of one of the synthetic samples from (Smith et al., 2018) left part of the scene is shown after texturing, while the right side is shown before texturing. b) A series of building in Unity engine[7]. c) synthetic dataset created in (Koch et al., 2019) using Blender.

Terragen[8], a scenery generator program, is another environment employed in the literature (Liu et al., 2019; Martin et al., 2015). Recently, the concept of an autonomous environment generator for UAV-based simulation based on machine learning algorithms is introduced in (Nakama et al., 2021). Based on satellite images, this approach introduced the concept of procedurally generating, scaling, and placing 3D models to create a realistic environment. The interested reader can refer to (Hentati et al., 2018; Nakama et al., 2021) for more details on simulation environments.

---

[1] https://www.unrealengine.com/en-US/

[2] https://vccimaging.org/Publications/Smith2018UAVPathPlanning/

[3] https://unity.com/

[4] https://microsoft.github.io/AirSim/

[5] https://3dwarehouse.sketchup.com/

[6] http://gazebosim.org

[7] https://unityassetsfreedom.club/download/town-constructor-pack/

[8] https://planetside.co.uk/

## 4   Discussion and future trends

Although much work and progress are observable in the field of viewpoint and path planning, existing algorithms are far from the last word on this topic, and there is plenty of exciting work left to do. In this section, we discuss the current challenges in viewpoint and path planning approaches which show the direction for future research on this topic.

Most of the model-free approaches rely on NBV and rapidly traverse along the direction that decreases the model uncertainty without using any prior information about the scene and object. Although this is very efficient for fast exploration of the objects, their local search strategy within a non-linear objective function ***may get stuck in local minima***. It cannot guarantee a complete coverage of the object, especially its details. Furthermore, most model-free methods use stereo cameras or depth cameras to capture data for 3D reconstruction. It means the platform should get close enough to the object. Although it is feasible for indoor applications and small objects, it could be very time-consuming or even impossible for large-scale 3D reconstruction purposes. On the other hand, having *a-priori* knowledge of the general geometry enables the explore-and-exploit methods to optimize the coverage and accuracy of the results globally leading to high-quality results and smoother trajectories. However, most of the existing explore-and-exploit approaches are off-line and do not update the computed poses based on the online feedback from the data capturing unit. Therefore, considering the current computational power of on-board processing units, refining the global design of the poses and updating them locally based on the feedback from the acquired images could be a very interesting direction for future research. Nowadays, this idea could also be more realistic with the availability of very fast data transfer technologies like 5G and the popularity of cloud-based processing. This also could close the gap between two flight missions (exploration and exploitation) which requires two visits from the site, which is not always possible in practice.

Like almost all other projects in engineering, one interesting direction could be ***as-planned vs. as-is*** investigation. Even in the case of a perfect planning, many other issues could affect the final outcome. Hardware limitations, meteorological conditions, objects complexities, and sensor-object relative conditions are the main factors that can greatly increase the uncertainty of the motion and state of the MAV or the quality of the final product. GNSS hardware and setup is an example in this regard which affects the data capturing quality. Although GNSS provides valuable information for UAV-based data capturing and processing, when the platforms fly at a low distance to the object of interest or under the structures like bridges, the GNSS signal can be strongly disturbed. Therefore, integrating visual SLAM and GNSS-based navigation and data capturing could be an interesting topic to have the best of both worlds. Drift on the gimbal and the magnetometer compass affects capturing the data with the already planned angles. Furthermore, due to safety issues e.g., minimum controllable flight height, some lower parts of the structures cannot be well captured and reconstructed (Kuang et al., 2020). To the best of our knowledge, most of the current approaches are designed for ideal conditions and neglect the effect of wind and other meteorological conditions. For example, wind affects the platform's ability to reach the planned viewpoints. Some studies (Kim and Eustice, 2015; Papachristos et al., 2019b) addressed the problems occurring by fast motion of the platform by considering the localization uncertainty of SLAM for view planning. They integrated exploration or coverage planning with an active SLAM approach, which aims to minimize the localization uncertainty of SLAM. They produced navigation paths tracking many SLAM features to reduce localization uncertainty. However, although their methods improved the localization accuracy, MAVs are still unable to navigate quickly because the generated paths do not satisfy MAV's dynamic properties. The integration of the active SLAM approaches (Kim and Eustice, 2015; Papachristos et al., 2019b) and the fast motion generation methods (Cieslewski et al., 2017; Dharmadhikari et al., 2020) could be another direction for future work. Since battery limitation (flight time) is a challenge for almost all systems, multi-agent systems like (He et al., 2021) seem to be very effective for large-scale 3D reconstruction purposes. However, current challenges like efficient and reliable tasks distribution between the agent are still challenging.

Besides all deviations caused by hardware imperfection and environmental conditions, the ***appearance of the objects with different materials*** can also affect the quality of 3D reconstruction. Shadow and illumination are also very important factors that can hinder reaching a high-fidelity 3D model. Reflective or low-texture surfaces like glass should also be considered. Moreover, since the texture is related to scale, it can pose another constraint on viewpoint planning which affects the acceptable distance between the sensor and low-texture parts of the objects. Hence, deviation of as-is poses of the captured images from as-planned ones necessitates an iterative closed-loop and online feedback between data capturing and planning modules that could be another direction for future research.

Since the perception of the surrounding environment is a key for collision-free UAV flight, commercial UAVs are equipped with some algorithms/sensors to perform a safe flight. However, when ***dealing with dynamic scenes*** (which is the case in many real scenarios), other considerations like moving objects detection also comes into play and should be considered in data capturing. Recently, (Tullu et al., 2021) employed a YoLO object detector (Redmon and Farhadi, 2018) to find obstacles and pedestrians in UAV's navigation environment. This would also be important to avoid any occlusion caused by moving objects between the sensor and the object of interest.

While some algorithms prioritize reconstruction quality, others emphasize the flight time (equivalently trajectory length or the number of images). Future research could also concentrate on finding reasonable and flexible solutions that can switch between these two possibilities or even add some hyperparameters for ***tuning the importance of 3D reconstruction quality and the trajectory length*** in order to fulfill the requirements of different applications.

*AI and Machine learning-based approaches:* Some studies recently applied machine learning techniques for viewpoint and path planning. Hepp et al. (2018a) proposed an NBV planning method where the viewpoint's utility is computed from a 3D convolutional neural network (CNN). The method used a multi-scale voxel representation of a partially explored scene as an input of the CNN. It trains utility scores of viewpoints from an oracle with access to ground truth information. Peralta et al. (Peralta et al., 2020) also proposed a learning-based NBV planning method for scanning houses. They provide a dataset of 3D house models for training NBV policies. The Authors trained a deep Q-network (Lillicrap et al., 2016) and deep deterministic policy gradient (Lillicrap et al., 2016) based on the house dataset. Zeng et al. (2020) proposed a deep-learning network for NBV planning, which directly uses raw point cloud data instead of a volumetric model. The network extracts the feature of partially reconstructed point cloud and predicts the information gain of viewpoints from the extracted feature. This network can estimate an information gain more efficiently than the conventional ray-casting approaches. An exploration planning method based on deep reinforcement learning is proposed in (Zhu et al., 2018). This method trains topological information of an office-like environment, which provides a guide to efficiently compute the visiting sequence for unexplored regions. Martin et al. (2016) proposed using genetic algorithms to optimize viewpoint planning for accurate 3D reconstruction using small UAVs. The coverage and accuracy of the generated model are formulated in the objective function of the Genetic algorithms. The objective function is defined based on the number of visible terrain points in the images of a solution set and the need to capture terrain points from multiple angles. The evaluation results show the improvement in the completeness and standard deviations of case studies in comparison with the basic grid survey. A hierarchical framework for path generation, coverage path planning, and dynamic obstacle avoidance is presented in (Lei et al., 2022). Satellite images and maps of farms were used via a deep learning method for path planning. A faster R-CNN network (Ren et al., 2017) localizes and identifies objects and obstacles of humans and vehicles. Xie et al. (2021) proposed a deep reinforcement learning approach for path planning in a dynamic environment. They formulated this issue as a Partially Observable Markov Decision Process and solved it by utilizing the local information and relative distance without global information. Historical state-action sequences and multi-sensors are used by the proposed method to achieve more reasonable decision-making. Although there is a rich body of literature on AI-based path and coverage planning (Aggarwal and Kumar, 2020; Kaba et al., 2017; Pehlivanoglu and Pehlivanoglu, 2021; Sonmez et al., 2015; Zhang et al., 2021; Zhao et al., 2018; Zhou et al., 2021), viewpoint planning for 3D reconstruction based on AI and machine/deep learning approaches are still in its infancy and there is a large room to deepen the insight into this topic.

**Acknowledgments**

This research is funded by the German Research Foundation (DFG) – TRR 277/1 2020 - Project number 414265976. M.Maboudi and M. Gerke would like to thank the DFG for the support within the SFB/Transregio 277- Additive manufacturing in construction. (Subproject C06).

**References**

Acharya, B.S., Bhandari, M., Bandini, F., Pizarro, A., Perks, M., Joshi, D.R., Wang, S., Dogwiler, T., Ray, R.L., Kharel, G., Sharma, S., 2021. Unmanned Aerial Vehicles in Hydrology and Water Management: Applications, Challenges, and Perspectives. Water Resour. Res. 57, e2021WR029925. https://doi.org/10.1029/2021WR029925

Aggarwal, S., Kumar, N., 2020. Path planning techniques for unmanned aerial vehicles: A review, solutions, and challenges. Comput. Commun. 149, 270–299. https://doi.org/10.1016/J.COMCOM.2019.10.014

Agisoft, 2022. Agisoft: Mission planning for complex structures [WWW Document]. URL https://agisoft.freshdesk.com/support/solutions/articles/31000157953-mission-planning-for-complex-structures (accessed 2.20.22).

Ahmadabadian, A.H., Robson, S., Boehm, J., Shortis, M., 2014. Stereo-imaging network design for precise and dense 3d reconstruction. Photogramm. Rec. 29, 317–336. https://doi.org/10.1111/phor.12076

Almadhoun, R., Taha, T., Seneviratne, L., Dias, J., Cai, G., 2016. A survey on inspecting structures using robotic systems: Int. J. Adv. Robot. Syst. 13, 1–18. https://doi.org/10.1177/1729881416663664

Almadhoun, R., Taha, T., Seneviratne, L., Zweiri, Y., 2019. A survey on multi-robot coverage path planning for model reconstruction and mapping. SN Appl. Sci. 2019 18 1, 1–24. https://doi.org/10.1007/S42452-019-0872-Y

Aloimonos, J., Weiss, I., Bandyopadhyay, A., 1988. Active vision. Int. J. Comput. Vis. 1987 14 1, 333–356. https://doi.org/10.1007/BF00133571

Alsadik, B., Remondino, F., 2020. Flight Planning for LiDAR-Based UAS Mapping Applications. ISPRS Int. J. Geo-Information 2020, Vol. 9, Page 378 9, 378. https://doi.org/10.3390/IJGI9060378

Arce Munoz, S., 2020. Optimized 3D Reconstruction for Infrastructure Inspection with Automated Structure from Motion and Machine Learning Methods Learning Methods. Brigham Young University.

Basiri, A., Mariani, V., Silano, G., Aatif, M., Iannelli, L., Glielmo, L., 2022. A survey on the application of path-planning algorithms for multi-rotor UAVs in precision agriculture. J. Navig. 1–20. https://doi.org/10.1017/S0373463321000825

Batinovic, A., Ivanovic, A., Petrovic, T., Bogdan, S., 2022. A Shadowcasting-Based Next-Best-View Planner for Autonomous 3D Exploration. IEEE Robot. Autom. Lett. 7, 2969–2976. https://doi.org/10.1109/LRA.2022.3146586

Batinovic, A., Petrovic, T., Ivanovic, A., Petric, F., Bogdan, S., 2021. A Multi-Resolution Frontier-Based Planner for Autonomous 3D Exploration. IEEE Robot. Autom. Lett. 6. https://doi.org/10.1109/LRA.2021.3068923

Bircher, A., Alexis, K., Burri, M., Oettershagen, P., Omari, S., Mantel, T., Siegwart, R., 2015. Structural inspection path planning via iterative viewpoint resampling with application to aerial robotics, in: International Conference on Robotics and Automation. IEEE, pp. 6423–6430. https://doi.org/10.1109/ICRA.2015.7140101

Bircher, A., Kamel, M., Alexis, K., Oleynikova, H., Siegwart, R., 2018. Receding horizon path planning for 3D exploration and surface inspection. Auton. Robots 42. https://doi.org/10.1007/s10514-016-9610-0

Bircher, A., Kamel, M., Alexis, K., Oleynikova, H., Siegwart, R., 2016. Receding horizon next-best-view planner for 3D exploration, in: Proceedings - IEEE International Conference on Robotics and Automation. https://doi.org/10.1109/ICRA.2016.7487281

Blender_Online_Community, 2018. Blender - a 3D modelling and rendering package. Stichting Blender Foundation, Amsterdam.

Bogaerts, B., Sels, S., Vanlanduit, S., Penne, R., 2019. Near-Optimal Path Planning for Complex Robotic Inspection Tasks 1–16.

Bolourian, N., Hammad, A., 2020. LiDAR-equipped UAV path planning considering potential locations of defects for bridge inspection. Autom. Constr. 117, 103250. https://doi.org/10.1016/J.AUTCON.2020.103250

Bouziani, M., Chaaba, H., Ettarid, M., 2021. Evaluation of 3D Building Model using Terrestrial Laser Scanning and Drone Photogrammetry. Int. Arch. Photogramm. Remote Sens. Spat. Inf. Sci. - ISPRS Arch. 46, 39–42. https://doi.org/10.5194/ISPRS-ARCHIVES-XLVI-4-W4-2021-39-2021

Bucker, A., Bonatti, R., Scherer, S., 2020. Do You See What I See? Coordinating Multiple Aerial Cameras for Robot Cinematography.

Cao, C., Zhu, H., Choset, H., Zhang, J., 2021. TARE: A Hierarchical Framework for Efficiently Exploring Complex 3D Environments. https://doi.org/10.15607/rss.2021.xvii.018

Charrow, B., Kahn, G., Patil, S., Liu, S., Goldberg, K., Abbeel, P., Michael, N., Kumar, V., 2015a. Information-theoretic planning with trajectory optimization for dense 3D mapping. Robot. Sci. Syst. 11. https://doi.org/10.15607/RSS.2015.XI.003

Charrow, B., Liu, S., Kumar, V., Michael, N., 2015b. Information-theoretic mapping using Cauchy-Schwarz Quadratic Mutual Information. Proc. - IEEE Int. Conf. Robot. Autom. 2015-June, 4791–4798. https://doi.org/10.1109/ICRA.2015.7139865

Chekuri, C., Pál, M., 2005. A recursive greedy algorithm for walks in directed graphs. Proc. - Annu. IEEE Symp. Found. Comput. Sci. FOCS 2005, 245–253. https://doi.org/10.1109/SFCS.2005.9

Chen, S., Li, Y., Kwok, N.M., 2011. Active vision in robotic systems: A survey of recent developments: Int. J. Rob. Res. 30, 1343–1377. https://doi.org/10.1177/0278364911410755

Chen, S.Y., Li, Y.F., 2005. Vision sensor planning for 3-D model acquisition. IEEE Trans. Syst. Man, Cybern. Part B Cybern. 35. https://doi.org/10.1109/TSMCB.2005.846907

Cieslewski, T., Kaufmann, E., Scaramuzza, D., 2017. Rapid exploration with multi-rotors: A frontier selection method for high speed flight, in: IEEE International Conference on Intelligent Robots and Systems. https://doi.org/10.1109/IROS.2017.8206030

Connolly, C.I., 1985. The determination of next best views. Proc. - IEEE Int. Conf. Robot. Autom. 432–435. https://doi.org/10.1109/ROBOT.1985.1087372

DadrasJavan, F., Samadzadegan, F., Seyed Pourazar, S.H., Fazeli, H., 2019. UAV-based multispectral imagery for fast Citrus Greening detection. J. Plant Dis. Prot. 2019 1264 126, 307–318. https://doi.org/10.1007/S41348-019-00234-8

Dai, A., Papatheodorou, S., Funk, N., Tzoumanikas, D., Leutenegger, S., 2020. Fast Frontier-based Information-driven Autonomous Exploration with an MAV, in: Proceedings - IEEE International Conference on Robotics and Automation. https://doi.org/10.1109/ICRA40945.2020.9196707

Daudelin, J., Campbell, M., 2017. An Adaptable, Probabilistic, Next-Best View Algorithm for Reconstruction of Unknown 3-D Objects. IEEE Robot. Autom. Lett. 2. https://doi.org/10.1109/LRA.2017.2660769

Debus, P., Rodehorst, V., 2021. Evaluation of 3D Uas Flight Path Planning Algorithms. Int. Arch. Photogramm. Remote Sens. Spat. Inf. Sci. XLIII-B1-2, 157–164. https://doi.org/10.5194/isprs-archives-xliii-b1-2021-157-2021

Dellaert, F., Kaess, M., 2006. Square root SAM: Simultaneous localization and mapping via square root information smoothing, in: International Journal of Robotics Research. https://doi.org/10.1177/0278364906072768

Delmerico, J., Isler, S., Sabzevari, R., Scaramuzza, D., 2018. A comparison of volumetric information gain metrics for active 3D object reconstruction. Auton. Robots 42. https://doi.org/10.1007/s10514-017-9634-0

Deris, A., Trigonis, I., Aravanis, A., Stathopoulou, E.K., 2017. Depth cameras on UAVs: A first approach, in: International Archives of the Photogrammetry, Remote Sensing and Spatial Information Sciences. International Society for Photogrammetry and Remote Sensing, pp. 231–236. https://doi.org/10.5194/ISPRS-ARCHIVES-XLII-2-W3-231-2017

Dharmadhikari, M., Dang, T., Solanka, L., Loje, J., Nguyen, H., Khedekar, N., Alexis, K., 2020. Motion Primitives-based Path Planning for Fast and Agile Exploration using Aerial Robots, in: Proceedings - IEEE International Conference on Robotics and Automation. https://doi.org/10.1109/ICRA40945.2020.9196964

Dissanayake, G., Sukkarieh, S., Nebot, E., Durrant-Whyte, H., 2001. The aiding of a low-cost strapdown inertial measurement unit using vehicle model constraints for land vehicle applications. IEEE Trans. Robot. Autom. 17. https://doi.org/10.1109/70.964672

Eltner, A., Hoffmeister, D., Kaiser, A., Karrasch, P., Klingbeil, L., Stöcker, C., Rovere, A., 2022. UAVs for the Environmental Sciences; Methods and Applications.

Eskandari, R., Mahdianpari, M., Mohammadimanesh, F., Salehi, B., Brisco, B., Homayouni, S., 2020. Meta-analysis of Unmanned Aerial Vehicle (UAV) Imagery for Agro-environmental Monitoring Using Machine Learning and Statistical Models. Remote Sens. 2020, Vol. 12, Page 3511 12, 3511. https://doi.org/10.3390/RS12213511

Estrada, C., Neira, J., Tardós, J.D., 2005. Hierarchical SLAM: Real-time accurate mapping of large environments. IEEE Trans. Robot. 21. https://doi.org/10.1109/TRO.2005.844673

Fang, B., Zhan, Z., 2020. A visual SLAM method based on point-line fusion in weak-matching scene: Int. J. Adv. Robot. Syst. 17. https://doi.org/10.1177/1729881420904193

Fiz, J.I., Martín, P.M., Cuesta, R., Subías, E., Codina, D., Cartes, A., 2022. Examples and Results of Aerial Photogrammetry in Archeology with UAV: Geometric Documentation, High Resolution Multispectral Analysis, Models and 3D Printing. Drones 2022, Vol. 6, Page 59 6, 59. https://doi.org/10.3390/DRONES6030059

Fraser, C., 1984. Network design considerations for non-topographic photogrammetry. Photogramm. Eng. Remote Sens. 50, 1115–1126.

Galceran, E., Carreras, M., 2013. A survey on coverage path planning for robotics. Rob. Auton. Syst. 61, 1258–1276. https://doi.org/10.1016/J.ROBOT.2013.09.004

Golden, B.L., Levy, L., Vohra, R., 1987. The orienteering problem. Nav. Res. Logist. 34, 307–318. https://doi.org/10.1002/1520-6750(198706)

González-Jorge, H., Martínez-Sánchez, J., Bueno, M., Arias, P., 2017. Unmanned Aerial Systems for Civil Applications: A Review. Drones 2017, Vol. 1, Page 2 1, 2. https://doi.org/10.3390/DRONES1010002

Gu, X., Fan, Z., Zhu, S., Dai, Z., Tan, F., Tan, P., 2020. Cascade Cost Volume for High-Resolution Multi-View Stereo and Stereo Matching, in: Proceedings of the IEEE Computer Society Conference on Computer Vision and Pattern Recognition. https://doi.org/10.1109/CVPR42600.2020.00257

Gunawan, A., Lau, H.C., Vansteenwegen, P., 2016. Orienteering Problem: A survey of recent variants, solution approaches and applications. Eur. J. Oper. Res. 255, 315–332. https://doi.org/10.1016/J.EJOR.2016.04.059

Hardouin, G., Morbidi, F., Moras, J., Marzat, J., Mouaddib, E.M., 2020. Surface-driven Next-Best-View planning for exploration of large-scale 3D environments, in: IFAC-PapersOnLine. https://doi.org/10.1016/j.ifacol.2020.12.2376

He, K., Gkioxari, G., Dollar, P., Girshick, R., 2017. Mask R-CNN, in: 2017 IEEE International Conference on Computer Vision (ICCV). IEEE, pp. 2980–2988. https://doi.org/10.1109/ICCV.2017.322

He, W., Qi, X., Liu, L., 2021. A novel hybrid particle swarm optimization for multi-UAV cooperate path planning. Appl. Intell. 51, 7350–7364. https://doi.org/10.1007/S10489-020-02082-8

Heng, L., Gotovos, A., Krause, A., Pollefeys, M., 2015. Efficient visual exploration and coverage with a micro aerial vehicle in unknown environments. Proc. - IEEE Int. Conf. Robot. Autom. 2015-June, 1071–1078. https://doi.org/10.1109/ICRA.2015.7139309

Hentati, A.I., Krichen, L., Fourati, M., Fourati, L.C., 2018. Simulation Tools, Environments and Frameworks for UAV Systems Performance Analysis. 2018 14th Int. Wirel. Commun. Mob. Comput. Conf. IWCMC 2018 1495–1500. https://doi.org/10.1109/IWCMC.2018.8450505

Hepp, B., 2018. Planning for Autonomous Micro-Aerial Vehicles with Applications to Filming and 3D Modeling. ETH Zürich.

Hepp, B., Dey, D., Sinha, S.N., Kapoor, A., Joshi, N., Hilliges, O., 2018a. Learn-to-score: Efficient 3D scene exploration by predicting view utility. Lect. Notes Comput. Sci. (including Subser. Lect. Notes Artif. Intell. Lect. Notes Bioinformatics) 11219 LNCS, 455–472. https://doi.org/10.1007/978-3-030-01267-0_27

Hepp, B., Niebner, M., Hilliges, O., 2018b. Plan3D: Viewpoint and trajectory optimization for aerial multi-view stereo reconstruction. ACM Trans. Graph. 38, 1–17. https://doi.org/10.1145/3233794

Hoppe, C., Wendel, A., Zollmann, S., Pirker, K., Irschara, A., Bischof, H., Kluckner, S., 2012. Photogrammetric Camera Network Design for Micro Aerial Vehicles Photogrammetric Camera Network Design for Micro Aerial Vehicles, in: 17th Computer Vision Winter Workshop.

Hornung, A., Wurm, K.M., Bennewitz, M., Stachniss, C., Burgard, W., 2013. OctoMap: An efficient probabilistic 3D mapping framework based on octrees. Auton. Robots 34. https://doi.org/10.1007/s10514-012-9321-0

Hornung, A., Zeng, B., Kobbelt, L., 2008. Image selection for improved multi-view stereo. 26th IEEE Conf. Comput. Vis. Pattern Recognition, CVPR. https://doi.org/10.1109/CVPR.2008.4587688

Hosseininaveh, A., Remondino, F., 2021. An imaging network design for UGV-based 3D reconstruction of buildings. Remote Sens. 13, 1–28. https://doi.org/10.3390/rs13101923

Huang, P., Lin, L., Xu, K., Huang, H., 2020. Autonomous Outdoor Scanning via Online Topological and Geometric Path Optimization. IEEE Trans. Intell. Transp. Syst. 1–14. https://doi.org/10.1109/TITS.2020.3039557

Huang, R., Zou, D., Vaughan, R., Tan, P., 2018. Active Image-Based Modeling with a Toy Drone, in: IEEE International Conference on Robotics and Automation. Institute of Electrical and Electronics Engineers Inc., pp. 6124–6131. https://doi.org/10.1109/ICRA.2018.8460673

Jacob-Loyola, N., Muñoz-La Rivera, F., Herrera, R.F., Atencio, E., 2021. Unmanned aerial vehicles (Uavs) for physical progress monitoring of construction. Sensors 21. https://doi.org/10.3390/S21124227

Jing, W., Deng, D., Wu, Y., Shimada, K., 2020. Multi-UAV coverage path planning for the inspection of large and complex structures. IEEE Int. Conf. Intell. Robot. Syst. 1480–1486. https://doi.org/10.1109/IROS45743.2020.9341089

Jing, W., Deng, D., Xiao, Z., Liu, Y., Shimada, K., 2019. Coverage Path Planning using Path Primitive Sampling and Primitive Coverage Graph for Visual Inspection. IEEE Int. Conf. Intell. Robot. Syst. 1472–1479. https://doi.org/10.1109/IROS40897.2019.8967969

Jing, W., Polden, J., Tao, P.Y., Lin, W., Shimada, K., 2016. View planning for 3D shape reconstruction of buildings with unmanned aerial vehicles, in: 14th International Conference on Control, Automation, Robotics and Vision, ICARCV 2016. IEEE. https://doi.org/10.1109/ICARCV.2016.7838774

Just, G.E., Pellenz, M.E., De Paula Lima, L.A., Chang, B.S., Souza, R.D., Montejo-Sánchez, S., 2020. UAV Path Optimization for Precision Agriculture Wireless Sensor Networks. Sensors 2020, Vol. 20, Page 6098 20, 6098. https://doi.org/10.3390/S20216098

Kaba, M.D., Uzunbas, M.G., Lim, S.N., 2017. A reinforcement learning approach to the view planning problem, in: 30th IEEE Conference on Computer Vision and Pattern Recognition, CVPR 2017. pp. 5094–5102. https://doi.org/10.1109/CVPR.2017.541

Karaman, S., Frazzoli, E., 2011. Sampling-based algorithms for optimal motion planning: http://dx.doi.org/10.1177/0278364911406761 30, 846–894. https://doi.org/10.1177/0278364911406761

Kazhdan, M., Hoppe, H., 2013. Screened poisson surface reconstruction. ACM Trans. Graph. 32. https://doi.org/10.1145/2487228.2487237

Kerle, N., Nex, F., Gerke, M., Duarte, D., Vetrivel, A., 2019. UAV-Based Structural Damage Mapping: A Review. ISPRS Int. J. Geo-Information 9, 14. https://doi.org/10.3390/ijgi9010014

Kim, A., Eustice, R.M., 2015. Active visual SLAM for robotic area coverage: Theory and experiment. Int. J. Rob. Res. 34. https://doi.org/10.1177/0278364914547893

Knapitsch, A., Park, J., Zhou, Q.-Y., Koltun, V., 2017. Tanks and Temples: Benchmarking Large-Scale Scene Reconstruction. ACM Trans. Graph. 36, 1–13. https://doi.org/10.1145/3072959.3073599

Koch, T., 2020. Automated and Precise 3D Building Reconstruction using UAVs. TU Munich.

Koch, T., Körner, M., Fraundorfer, F., 2019. Automatic and Semantically-Aware 3D UAV Flight Planning for Image-Based 3D Reconstruction. Remote Sens. 11, 1550. https://doi.org/10.3390/rs11131550

Kompis, Y., Bartolomei, L., Mascaro, R., Teixeira, L., Chli, M., 2021. Informed Sampling Exploration Path Planner for 3D Reconstruction of Large Scenes. https://doi.org/10.3929/ethz-b-000501333

Kuang, Q., Wu, J., Pan, J., Zhou, B., 2020. Real-Time UAV Path Planning for Autonomous Urban Scene Reconstruction, in: International Conference on Robotics and Automation. IEEE, pp. 1156–1162. https://doi.org/10.1109/ICRA40945.2020.9196558

Lee, I.D., Seo, J.H., Kim, Y.M., Choi, J., Han, S., Yoo, B., 2020. Automatic Pose Generation for Robotic 3-D Scanning of Mechanical Parts. IEEE Trans. Robot. 36, 1219–1238. https://doi.org/10.1109/TRO.2020.2980161

Lei, T., Luo, C., Jan, G.E., Bi, Z., 2022. Deep Learning-Based Complete Coverage Path Planning With Re-Joint and Obstacle Fusion Paradigm. Front. Robot. AI 9, 1–14. https://doi.org/10.3389/frobt.2022.843816

Leonard, J.J., Durrant-Whyte, H.F., 1991. Simultaneous map building and localization for an autonomous mobile robot, in: IROS '91:IEEE/RSJ International Workshop on Intelligent Robots and Systems. IEEE, pp. 1442–1447. https://doi.org/10.1109/IROS.1991.174711

Li, Y., Liu, C., 2018. Applications of multirotor drone technologies in construction management. Int. J. Constr. Manag. 19, 401–412. https://doi.org/10.1080/15623599.2018.1452101

Lillicrap, T.P., Hunt, J.J., Pritzel, A., Heess, N., Erez, T., Tassa, Y., Silver, D., Wierstra, D., 2016. Continuous control with deep reinforcement learning, in: 4th International Conference on Learning Representations, ICLR 2016 - Conference Track Proceedings.

Liu, C., Zhang, S., Akbar, A., 2019. Ground Feature Oriented Path Planning for Unmanned Aerial Vehicle Mapping. IEEE J. Sel. Top. Appl. Earth Obs. Remote Sens. 12, 1175–1187. https://doi.org/10.1109/JSTARS.2019.2899369

Long, J., Shelhamer, E., Darrell, T., 2015. Fully convolutional networks for semantic segmentation. Proc. IEEE Comput. Soc. Conf. Comput. Vis. Pattern Recognit. 07-12-June-2015, 431–440. https://doi.org/10.1109/CVPR.2015.7298965

Maboudi, M., Alamouri, A., López De Arriba, V., Bajauri, M.S., Berger, C., Gerke, M., 2021. Drone-Based Container Crane Inspection: Concept, Challenges and Preliminary Results, in: ISPRS Annals of the Photogrammetry, Remote Sensing and Spatial Information Sciences. pp. 121–128. https://doi.org/10.5194/ISPRS-ANNALS-V-1-2021-121-2021

Maboudi, M., Bánhidi, D., Gerke, M., 2018. Investigation of Geometric Performance of an Indoor Mobile Mapping System, in: ISPRS - International Archives of the Photogrammetry, Remote Sensing and Spatial Information Sciences. pp. 637–642. https://doi.org/10.5194/isprs-archives-XLII-2-637-2018

Maldonado, O.A.M., Hadfield, S., Pugeault, N., Bowden, R., 2016. Next-Best Stereo: Extending Next-Best View Optimisation For Collaborative Sensors, in: Proceedings of the British Machine Vision Conference (BMVC). pp. 65.1-65.12. https://doi.org/10.5244/C.30.65

Malihi, S., Zoej, M.J.V., Hahn, M., Mokhtarzade, M., 2018. Window Detection from UAS-Derived Photogrammetric Point Cloud Employing Density-Based Filtering and Perceptual Organization. Remote Sens. 2018, Vol. 10, Page 1320 10, 1320. https://doi.org/10.3390/RS10081320

Martin, R.A., Rojas, I., Franke, K., Hedengren, J.D., 2015. Evolutionary View Planning for Optimized UAV Terrain Modeling in a Simulated Environment. Remote Sens. 2016, Vol. 8, Page 26 8, 26. https://doi.org/10.3390/RS8010026

Mason, S., 1997. Heuristic Reasoning Strategy for Automated Sensor Placement. Photogramm. Eng. Remote Sens. 63, 1093–1102.

Mauro, M., Riemenschneider, H., Signoroni, A., Leonardi, R., Van Gool, L., 2014. A unified framework for content-aware view selection and planning through view importance. Proc. BMVC 2014 1–11.

Meng, Z., Qin, H., Chen, Z., Chen, X., Sun, H., Lin, F., Ang, M.H., 2017. A two-stage optimized next-view planning framework for 3-D unknown environment exploration, and structural reconstruction. IEEE Robot. Autom. Lett. 2. https://doi.org/10.1109/LRA.2017.2655144

Monica, R., Aleotti, J., 2018. Contour-based next-best view planning from point cloud segmentation of unknown objects. Auton. Robots 42. https://doi.org/10.1007/s10514-017-9618-0

Montemerlo, M., Thrun, S., Koller, D., Wegbreit, B., 2002. FastSLAM: A factored solution to the simultaneous localization and mapping problem, in: Proceedings of the National Conference on Artificial Intelligence.

Moravec, H.P., Elfes, A., 1985. High resolution maps from wide angle sonar, in: Proceedings - IEEE International Conference on Robotics and Automation. https://doi.org/10.1109/ROBOT.1985.1087316

Mostegel, C., Rumpler, M., Fraundorfer, F., Bischof, H., 2016. UAV-Based Autonomous Image Acquisition with Multi-view Stereo Quality Assurance by Confidence Prediction, in: IEEE Computer Society Conference on Computer Vision and Pattern Recognition Workshops. IEEE Computer Society, pp. 1–10. https://doi.org/10.1109/CVPRW.2016.8

Mur-Artal, R., Montiel, J.M.M., Tardos, J.D., 2015. ORB-SLAM: A Versatile and Accurate Monocular SLAM System. IEEE Trans. Robot. 31. https://doi.org/10.1109/TRO.2015.2463671

Nagasawa, R., Mas, E., Moya, L., Koshimura, S., 2021. Model-based analysis of multi-UAV path planning for surveying postdisaster building damage. Sci. Reports 2021 111 11, 1–14. https://doi.org/10.1038/s41598-021-97804-4

Nakama, J., Parada, R., Matos-Carvalho, J.P., Azevedo, F., Pedro, D., Campos, L., 2021. Autonomous Environment Generator for UAV-Based Simulation. Appl. Sci. 11, 2185. https://doi.org/10.3390/app11052185

Newcombe, R.A., Izadi, S., Hilliges, O., Molyneaux, D., Kim, D., Davison, A.J., Kohli, P., Shotton, J., Hodges, S., Fitzgibbon, A., 2011. KinectFusion: Real-time dense surface mapping and tracking, in: 2011 10th IEEE International Symposium on Mixed and Augmented Reality, ISMAR 2011. https://doi.org/10.1109/ISMAR.2011.6092378

Nex, F., Armenakis, C., Cramer, M., Cucci, D.A., Gerke, M., Honkavaara, E., Kukko, A., Persello, C., Skaloud, J., 2022. UAV in the advent of the twenties: Where we stand and what is next. ISPRS J. Photogramm. Remote Sens. 184, 215–242. https://doi.org/10.1016/J.ISPRSJPRS.2021.12.006

Oleynikova, H., Taylor, Z., Fehr, M., Siegwart, R., Nieto, J., 2017. Voxblox: Incremental 3D Euclidean Signed Distance Fields for on-board MAV planning, in: IEEE International Conference on Intelligent Robots and Systems. https://doi.org/10.1109/IROS.2017.8202315

Oßwald, S., Bennewitz, M., Burgard, W., Stachniss, C., 2016. Speeding-Up Robot Exploration by Exploiting Background Information. IEEE Robot. Autom. Lett. 1. https://doi.org/10.1109/LRA.2016.2520560

Panda, M., Das, B., Subudhi, B., Pati, B.B., 2020. A Comprehensive Review of Path Planning Algorithms for Autonomous Underwater Vehicles. Int. J. Autom. Comput. 2019 173 17, 321–352. https://doi.org/10.1007/S11633-019-1204-9

Papachristos, C., Kamel, M., Popović, M., Khattak, S., Bircher, A., Oleynikova, H., Dang, T., Mascarich, F., Alexis, K., Siegwart, R., 2019a. Autonomous Exploration and Inspection Path Planning for Aerial Robots Using the Robot Operating System. Stud. Comput. Intell. 778, 67–111. https://doi.org/10.1007/978-3-319-91590-6_3

Papachristos, C., Mascarich, F., Khattak, S., Dang, T., Alexis, K., 2019b. Localization uncertainty-aware autonomous exploration and mapping with aerial robots using receding horizon path-planning. Auton. Robots 43. https://doi.org/10.1007/s10514-019-09864-1

Pehlivanoglu, Y.V., Pehlivanoglu, P., 2021. An enhanced genetic algorithm for path planning of autonomous UAV in target coverage problems. Appl. Soft Comput. 112, 107796. https://doi.org/https://doi.org/10.1016/j.asoc.2021.107796

Peng, C., Isler, V., 2020. Visual coverage path planning for urban environments. IEEE Robot. Autom. Lett. 5, 5961–5968. https://doi.org/10.1109/LRA.2020.3010745

Peng, C., Isler, V., 2019. Adaptive View Planning for Aerial 3D Reconstruction, in: Proceedings - IEEE International Conference on Robotics and Automation. Institute of Electrical and Electronics Engineers Inc., pp. 2981–2987.

Peng, C., Isler, V., 2018. View Selection with Geometric Uncertainty Modelling, in: Robotics: Science and Systems. https://doi.org/10.15607/RSS.2018.XIV.025

Peng, C., Isler, V., 2017. Optimal Reconstruction with a Small Number of Views. CoRR abs/1704.0.

Peralta, D., Casimiro, J., Nilles, A.M., Aguilar, J.A., Atienza, R., Cajote, R., 2020. Next-Best View Policy for 3D Reconstruction, in: Lecture Notes in Computer Science (Including Subseries Lecture Notes in Artificial Intelligence and Lecture Notes in Bioinformatics). https://doi.org/10.1007/978-3-030-66823-5_33

Pizzoli, M., Forster, C., Scaramuzza, D., 2014. REMODE: Probabilistic, monocular dense reconstruction in real time, in: Proceedings - IEEE International Conference on Robotics and Automation. https://doi.org/10.1109/ICRA.2014.6907233

Popovic, M., Hitz, G., Nieto, J., Sa, I., Siegwart, R., Galceran, E., 2017. Online informative path planning for active classification using UAVs. Proc. - IEEE Int. Conf. Robot. Autom. 5753–5758. https://doi.org/10.1109/ICRA.2017.7989676

Qin, H., Meng, Z., Meng, W., Chen, X., Sun, H., Lin, F., Ang, M.H., 2019. Autonomous Exploration and Mapping System Using Heterogeneous UAVs and UGVs in GPS-Denied Environments. IEEE Trans. Veh. Technol. 68, 1339–1350. https://doi.org/10.1109/TVT.2018.2890416

Qiu, W., Zhong, F., Zhang, Y., Qiao, S., Xiao, Z., Kim, T.S., Wang, Y., Yuille, A., 2017. UnrealCV: Virtual worlds for computer vision. MM 2017 - Proc. 2017 ACM Multimed. Conf. 1221–1224. https://doi.org/10.1145/3123266.3129396

Radoglou-Grammatikis, P., Sarigiannidis, P., Lagkas, T., Moscholios, I., 2020. A compilation of UAV applications for precision agriculture. Comput. Networks 172, 107148. https://doi.org/10.1016/J.COMNET.2020.107148

Rakha, T., Gorodetsky, A., 2018. Review of Unmanned Aerial System (UAS) applications in the built environment: Towards automated building inspection procedures using drones. Autom. Constr. 93, 252–264. https://doi.org/10.1016/J.AUTCON.2018.05.002

Redmon, J., Farhadi, A., 2018. YOLOv3: An Incremental Improvement. https://doi.org/10.48550/arxiv.1804.02767

Ren, S., He, K., Girshick, R., Sun, J., 2017. Faster R-CNN: Towards Real-Time Object Detection with Region Proposal Networks. IEEE Trans. Pattern Anal. Mach. Intell. 39, 1137–1149. https://doi.org/10.1109/TPAMI.2016.2577031

Roberts, M., 2019. Trajectory optimization methods for drone cameras.

Roberts, M., Shah, S., Dey, D., Truong, A., Sinha, S., Kapoor, A., Hanrahan, P., Joshi, N., 2017. Submodular Trajectory Optimization for Aerial 3D Scanning, in: Proceedings of the IEEE International Conference on Computer Vision. IEEE, pp. 5334–5343. https://doi.org/10.1109/ICCV.2017.569

Rottensteiner, F., Sohn, G., Gerke, M., Wegner, J.D., Breitkopf, U., Jung, J., 2014. Results of the ISPRS benchmark on urban object detection and 3D building reconstruction. ISPRS J. Photogramm. Remote Sens. 93, 256–271. https://doi.org/10.1016/J.ISPRSJPRS.2013.10.004

Saadatseresht, M., Samadzadegan, F., Azizi, A., 2005. Automatic camera placement in vision metrology based on a fuzzy inference system. Photogramm. Eng. Remote Sensing 71, 1375–1385. https://doi.org/10.14358/PERS.71.12.1375

Sadeghi, A., Asghar, A.B., Smith, S.L., 2019. On Minimum Time Multi-Robot Planning with Guarantees on the Total Collected Reward, in: International Symposium on Multi-Robot and Multi-Agent Systems, MRS 2019. Institute of Electrical and Electronics Engineers Inc., pp. 16–22. https://doi.org/10.1109/MRS.2019.8901079

Sahebdivani, S., Arefi, H., Maboudi, M., 2020. Rail Track Detection and Projection-Based 3D Modeling from UAV Point Cloud. Sensors 20, 5220. https://doi.org/10.3390/s20185220

Saponaro, M., Capolupo, A., Tarantino, E., Fratino, U., 2019. Comparative Analysis of Different UAV-Based Photogrammetric Processes to Improve Product Accuracies. Lect. Notes Comput. Sci. (including Subser. Lect. Notes Artif. Intell. Lect. Notes Bioinformatics) 11622 LNCS, 225–238. https://doi.org/10.1007/978-3-030-24305-0_18

Scaramuzza, D., Achtelik, M.C., Doitsidis, L., Friedrich, F., Kosmatopoulos, E., Martinelli, A., Achtelik, M.W., Chli, M., Chatzichristofis, S., Kneip, L., Gurdan, D., Heng, L., Lee, G.H., Lynen, S., Pollefeys, M., Renzaglia, A., Siegwart, R., Stumpf, J.C., Tanskanen, P., Troiani, C., Weiss, S., Meier, L., 2014. Vision-controlled micro flying robots: From system design to autonomous navigation and mapping in GPS-denied environments. IEEE Robot. Autom. Mag. 21, 26–40. https://doi.org/10.1109/MRA.2014.2322295

Schmid, K., Hirschmüller, H., Dömel, A., Grixa, I., Suppa, M., Hirzinger, G., 2012. View planning for multi-view stereo 3D Reconstruction using an autonomous multicopter. J. Intell. Robot. Syst. Theory Appl. 65, 309–323. https://doi.org/10.1007/s10846-011-9576-2

Schmid, L., Pantic, M., Khanna, R., Ott, L., Siegwart, R., Nieto, J., 2020. An Efficient Sampling-Based Method for Online Informative Path Planning in Unknown Environments. IEEE Robot. Autom. Lett. 5. https://doi.org/10.1109/LRA.2020.2969191

Schönberger, J.L., Zheng, E., Frahm, J.M., Pollefeys, M., 2016. Pixelwise View Selection for Unstructured Multi-View Stereo. Lect. Notes Comput. Sci. (including Subser. Lect. Notes Artif. Intell. Lect. Notes Bioinformatics) 9907 LNCS, 501–518. https://doi.org/10.1007/978-3-319-46487-9_31

Scott, W.R., Roth, G., Rivest, J.F., 2003. View planning for automated three-dimensional object reconstruction and inspection. ACM Comput. Surv. 35, 64–96. https://doi.org/10.1145/641865.641868

Seitz, S.M., Curless, B., Diebel, J., Scharstein, D., Szeliski, R., 2006. A comparison and evaluation of multi-view stereo reconstruction algorithms. Proc. IEEE Comput. Soc. Conf. Comput. Vis. Pattern Recognit. 1, 519–526. https://doi.org/10.1109/CVPR.2006.19

Seo, J., Duque, L., Wacker, J., 2018. Drone-enabled bridge inspection methodology and application. Autom. Constr. 94, 112–126. https://doi.org/10.1016/J.AUTCON.2018.06.006

Shade, R., Newman, P., 2011. Choosing where to go: Complete 3D exploration with Stereo, in: Proceedings - IEEE International Conference on Robotics and Automation. https://doi.org/10.1109/ICRA.2011.5980121

Shah, S., Dey, D., Lovett, C., Kapoor, A., 2017. AirSim: High-Fidelity Visual and Physical Simulation for Autonomous Vehicles, in: Field and Service Robotics Conference (FSR 2017). Springer Science and Business Media B.V., pp. 621–635. https://doi.org/10.48550/arxiv.1705.05065

Shang, Z., Bradley, J., Shen, Z., 2020. A co-optimal coverage path planning method for aerial scanning of complex structures. Expert Syst. Appl. 158, 113535. https://doi.org/10.1016/J.ESWA.2020.113535

Sharma, O., Arora, N., Sagar, H., 2019. Image Acquisition for High Quality Architectural Reconstruction, in: Graphics Interface 2019. Kingston, Ontario. https://doi.org/10.20380/GI2019.18

Shen, S., Michael, N., Kumar, V., 2012. Autonomous indoor 3D exploration with a micro-aerial vehicle. Proc. - IEEE Int. Conf. Robot. Autom. 9–15. https://doi.org/10.1109/ICRA.2012.6225146

Smith, N., Moehrle, N., Goesele, M., Heidrich, W., 2018. Aerial path planning for urban scene reconstruction: A continuous optimization method and benchmark, in: SIGGRAPH Asia 2018 Technical Papers, SIGGRAPH Asia 2018. https://doi.org/10.1145/3272127.3275010

Song, S., Jo, S., 2018. Surface-Based Exploration for Autonomous 3D Modeling. Proc. - IEEE Int. Conf. Robot. Autom. 4319–4326. https://doi.org/10.1109/ICRA.2018.8460862

Song, S., Jo, S., 2017. Online inspection path planning for autonomous 3D modeling using a micro-aerial vehicle, in: IEEE International Conference on Robotics and Automation (ICRA). https://doi.org/10.1109/ICRA.2017.7989737

Song, S., Kim, D., Choi, S., 2021. View Path Planning via Online Multiview Stereo for 3-D Modeling of Large-Scale Structures. IEEE Trans. Robot. 38, 372–390. https://doi.org/10.1109/tro.2021.3083197

Song, S., Kim, D., Jo, S., 2020a. Online coverage and inspection planning for 3D modeling. Auton. Robot. 2020 448 44, 1431–1450. https://doi.org/10.1007/S10514-020-09936-7

Song, S., Kim, D., Jo, S., 2020b. Active 3D Modeling via Online Multi-View Stereo. Proc. - IEEE Int. Conf. Robot. Autom. 5284–5291. https://doi.org/10.1109/ICRA40945.2020.9197089

Sonmez, A., Kocyigit, E., Kugu, E., 2015. Optimal path planning for UAVs using Genetic Algorithm. 2015 Int. Conf. Unmanned Aircr. Syst. ICUAS 2015 50–55. https://doi.org/10.1109/ICUAS.2015.7152274

Stache, F., Westheider, J., Magistri, F., Popovic, M., Stachniss, C., 2021. Adaptive path planning for UAV-based multi-resolution semantic segmentation. 2021 10th Eur. Conf. Mob. Robot. ECMR 2021 - Proc. https://doi.org/10.1109/ECMR50962.2021.9568788

Sun, Y., Huang, Q., Hsiao, D.-Y., Guan, L., Hua, G., 2021. Learning View Selection for 3D Scenes. Proc. IEEE/CVF Conf. Comput. Vis. Pattern Recognit. 14464–14473.

Sun, Y., Ma, O., 2022. Automating Aircraft Scanning for Inspection or 3D Model Creation with a UAV and Optimal Path Planning. Drones 2022, Vol. 6, Page 87 6, 87. https://doi.org/10.3390/DRONES6040087

Tan, C.S., Mohd-Mokhtar, R., Arshad, M.R., 2021. A Comprehensive Review of Coverage Path Planning in Robotics Using Classical and Heuristic Algorithms. IEEE Access 9, 119310–119342. https://doi.org/10.1109/ACCESS.2021.3108177

Torresan, C., Berton, A., Carotenuto, F., Di Gennaro, S.F., Gioli, B., Matese, A., Miglietta, F., Vagnoli, C., Zaldei, A., Wallace, L., 2016. Forestry applications of UAVs in Europe: a review. Int. J. Remote Sens. 38, 2427–2447. https://doi.org/10.1080/01431161.2016.1252477

Tsagris, M., Beneki, C., Hassani, H., 2014. On the Folded Normal Distribution. Math. 2014, Vol. 2, Pages 12-28 2, 12–28. https://doi.org/10.3390/MATH2010012

Tullu, A., Endale, B., Wondosen, A., Hwang, H.-Y., 2021. Machine Learning Approach to Real-Time 3D Path Planning for Autonomous Navigation of Unmanned Aerial Vehicle. Appl. Sci. 11, 4706. https://doi.org/10.3390/APP11104706

Ułanowicz, L., Sabak, R., 2021. Unmanned aerial vehicles supporting imagery intelligence using the structured light technology. Arch. Transp. 58, 35–45. https://doi.org/10.5604/01.3001.0014.8796

Umari, H., Mukhopadhyay, S., 2017. Autonomous robotic exploration based on multiple rapidly-exploring randomized trees, in: IEEE International Conference on Intelligent Robots and Systems. https://doi.org/10.1109/IROS.2017.8202319

Valente, J., Cerro, J. Del, Barrientos, A., Sanz, D., 2013. Aerial coverage optimization in precision agriculture management: A musical harmony inspired approach. Comput. Electron. Agric. 99, 153–159. https://doi.org/10.1016/J.COMPAG.2013.09.008

Vansteenwegen, P., Souffriau, W., Oudheusden, D. Van, 2011. The orienteering problem: A survey. Eur. J. Oper. Res. 209, 1–10. https://doi.org/10.1016/J.EJOR.2010.03.045

Vasquez-Gomez, J.I., Lopez-Damian, E., Sucar, L.E., 2009. View planning for 3D object reconstruction, in: IEEE/RSJ International Conference on Intelligent Robots and Systems, IROS 2009. pp. 4015–4020. https://doi.org/10.1109/IROS.2009.5354383

Vasquez-Gomez, J.I., Sucar, L.E., Murrieta-Cid, R., 2017. View/state planning for three-dimensional object reconstruction under uncertainty. Auton. Robots 41. https://doi.org/10.1007/s10514-015-9531-3

Wang, C., Ma, H., Chen, W., Liu, L., Meng, M.Q.H., 2020. Efficient Autonomous Exploration with Incrementally Built Topological Map in 3-D Environments. IEEE Trans. Instrum. Meas. 69, 9853–9865. https://doi.org/10.1109/TIM.2020.3001816

Whelan, T., Kaess, M., Johannsson, H., Fallon, M., Leonard, J.J., McDonald, J., 2014. Real-time large-scale dense

RGB-D SLAM with volumetric fusion: Int. J. Rob. Res. 34, 598–626. https://doi.org/10.1177/0278364914551008

Whelan, T., Salas-Moreno, R.F., Glocker, B., Davison, A.J., Leutenegger, S., 2016. ElasticFusion: Real-time dense SLAM and light source estimation, in: International Journal of Robotics Research. https://doi.org/10.1177/0278364916669237

Wu, S., Sun, W., Long, P., Huang, H., Cohen-Or Daniel, Gong, M., Deussen, O., Chen, B., 2014. Quality-driven poisson-guided autoscanning. ACM Trans. Graph. 33, 203. https://doi.org/10.1145/2661229.2661242

Xie, R., Meng, Z., Wang, L., Li, H., Wang, K., Wu, Z., 2021. Unmanned Aerial Vehicle Path Planning Algorithm Based on Deep Reinforcement Learning in Large-Scale and Dynamic Environments. IEEE Access 9. https://doi.org/10.1109/ACCESS.2021.3057485

Xu, L., Cheng, W., Guo, K., Han, L., Liu, Y., Fang, L., 2021. FlyFusion: Realtime Dynamic Scene Reconstruction Using a Flying Depth Camera. IEEE Trans. Vis. Comput. Graph. 27, 68–82. https://doi.org/10.1109/TVCG.2019.2930691

Xu, Y., Turkan, Y., 2020. BrIM and UAS for bridge inspections and management. Eng. Constr. Archit. Manag. 27, 785–807. https://doi.org/10.1108/ECAM-12-2018-0556/FULL/XML

Yamauchi, B., 1997. Frontier-based approach for autonomous exploration, in: Proceedings of IEEE International Symposium on Computational Intelligence in Robotics and Automation, CIRA. https://doi.org/10.1109/cira.1997.613851

Yan, F., Xia, E., Li, Z., Zhou, Z., 2021. Sampling-based path planning for high-quality aerial 3D reconstruction of urban scenes. Remote Sens. 13, 1–23. https://doi.org/10.3390/rs13050989

Yoder, L., Scherer, S., 2016. Autonomous Exploration for Infrastructure Modeling with a Micro Aerial Vehicle, in: Wettergreen, D.S., Barfoot, T.D. (Eds.), Field and Service Robotics: Results of the 10th International Conference. Springer International Publishing, Cham, pp. 427–440. https://doi.org/10.1007/978-3-319-27702-8_28

Zeng, R., Zhao, W., Liu, Y.J., 2020. PC-NBV: A point cloud based deep network for efficient next best view planning, in: IEEE International Conference on Intelligent Robots and Systems. https://doi.org/10.1109/IROS45743.2020.9340916

Zhang, S., Liu, C., Haala, N., 2020. Three-Dimensional Path Planning of Uavs Imaging for Complete Photogrammetric Reconstruction, in: ISPRS Annals of the Photogrammetry, Remote Sensing and Spatial Information Sciences. https://doi.org/10.5194/isprs-annals-V-1-2020-325-2020

Zhang, W., Zhang, S., Wu, F., Wang, Y., 2021. Path Planning of UAV Based on Improved Adaptive Grey Wolf Optimization Algorithm. IEEE Access 9, 89400–89411. https://doi.org/10.1109/ACCESS.2021.3090776

Zhao, Y., Zheng, Z., Liu, Y., 2018. Survey on computational-intelligence-based UAV path planning. Knowledge-Based Syst. 158, 54–64. https://doi.org/10.1016/J.KNOSYS.2018.05.033

Zhou, X., Gao, F., Fang, X., Lan, Z., 2021. Improved Bat Algorithm for UAV Path Planning in Three-Dimensional Space. IEEE Access 9, 20100–20116. https://doi.org/10.1109/ACCESS.2021.3054179

Zhou, X., Xie, K., Huang, K., Liu, Y., Zhou, Y., Gong, M., Huang, H., 2020a. Offsite aerial path planning for efficient urban scene reconstruction. ACM Trans. Graph. 39. https://doi.org/10.1145/3414685.3417791

Zhou, X., Yi, Z., Liu, Y., Huang, K., Huang, H., 2020b. Survey on path and view planning for UAVs. Virtual Real. Intell. Hardw. 2, 56–69. https://doi.org/10.1016/j.vrih.2019.12.004

Zhu, D., Li, T., Ho, D., Wang, C., Meng, M.Q.H., 2018. Deep Reinforcement Learning Supervised Autonomous Exploration in Office Environments, in: Proceedings - IEEE International Conference on Robotics and Automation. https://doi.org/10.1109/ICRA.2018.8463213